\documentclass{bmvc2k}

\title{\tool{}: Downstream Models are Vulnerable to Blackbox Attacks by 3D Gaussian Splat Camouflages}

\addauthor{Matthew Hull, Haoyang Yang, Pratham Mehta, Mansi Phute, Aeree Cho, Haoran Wang, Matthew Lau, Wenke Lee}{matthewhull@gatech.edu}{1}
\addauthor{Willian Lunardi}{williantessaro.lunardi@tii.ae}{2}
\addauthor{Martin Andreoni}{martin.andreoni@ku.ac.ae}{3}
\addauthor{Duen Horng Chau}{polo@gatech.edu}{1}
\addinstitution{
  Georgia Institute of Technology\\
  Atlanta, GA, USA
}
\addinstitution{
  Technology Innovation Institute\\
  Abu Dhabi, UAE
}
\addinstitution{
  Khalifa University\\
  Abu Dhabi, UAE
}

\runninghead{Hull \bmvaEtAl}{ComplicitSplat}

\usepackage{graphicx}
\usepackage{caption}
\usepackage{booktabs}
\usepackage{float}
\usepackage{array}
\usepackage{enumitem}
\usepackage{tikz}
\usepackage{fontawesome5}
\usepackage{ifthen}
\usepackage{xspace}
\usepackage{multirow}
\usepackage{xcolor}
\usepackage{siunitx}
\usepackage[accsupp]{axessibility}
\usepackage{hyperref}

\definecolor{agreen}{RGB}{74, 198, 148}
\definecolor{purple}{RGB}{158, 62, 177}
\definecolor{darkpurple}{RGB}{170, 70, 210}
\definecolor{aqua}{RGB}{87, 180, 181}
\definecolor{lightblue}{RGB}{72, 123, 232}
\definecolor{hotpink}{RGB}{255, 83, 115}
\definecolor{teal}{RGB}{90, 200, 250}
\definecolor{linkColor}{RGB}{0, 128, 229}
\definecolor{lightgreen}{RGB}{33, 222, 128}
\definecolor{almostBlack}{RGB}{60,60,60}

\definecolor{red}{RGB}{236, 107, 44}
\definecolor{green}{RGB}{0, 128, 0}
\definecolor{yellow}{RGB}{255, 192, 0}
\definecolor{purple}{RGB}{128, 0, 128}
\definecolor{cyan}{RGB}{0, 255, 255}
\definecolor{lightgray}{gray}{0.95}
\definecolor{grayborder}{gray}{0.5}
\definecolor{gray}{gray}{0.75}
\definecolor{orange}{RGB}{236, 107, 44}
\definecolor{lightorange}{RGB}{255, 223, 186}
\definecolor{blue}{RGB}{116, 95, 232}
\definecolor{lightblue}{RGB}{255, 223, 186}
\definecolor{lightpurple}{RGB}{202,58,126}

\definecolor{9colorq1}{RGB}{166,206,227}
\definecolor{9colorq2}{RGB}{31,120,180}
\definecolor{9colorq3}{RGB}{178,223,138}
\definecolor{9colorq4}{RGB}{51,160,44}
\definecolor{9colorq5}{RGB}{251,154,153}
\definecolor{9colorq6}{RGB}{227,26,28}
\definecolor{9colorq7}{RGB}{253,191,111}
\definecolor{9colorq8}{RGB}{255,127,0}
\definecolor{9colorq9}{RGB}{202,178,214}

\newcommand{\tool}{ComplicitSplat}

\newcommand{\tagOne}[1]{\ifthenelse{\equal{#1}{cap}}{Viewpoint-Specific Camouflage}{viewpoint-specific camouflage}\xspace}
\newcommand{\tagTwo}[1]{\ifthenelse{\equal{#1}{cap}}{Generalizes Across Detectors}{generalizes across detectors}\xspace}
\newcommand{\tagThree}[1]{\ifthenelse{\equal{#1}{cap}}{Cross-Domain Attack}{cross-domain attack}\xspace}

\newcommand{\CarBlue}   {{\color{blue}\faCar}}
\newcommand{\CarRed}    {{\color{red}\faCar}}
\newcommand{\CarGray}   {{\color{darkgray}\faCar}}

\newcommand{\TikZStop}{%
  \tikz[baseline=-0.75ex, scale=0.7, rotate=22.5]{%
    \draw[fill=red,draw=none]
    (0:0.15) -- (45:0.15) -- (90:0.15) -- (135:0.15)
    -- (180:0.15) -- (225:0.15) -- (270:0.15) -- (315:0.15) -- cycle;
  }%
}

\newcommand{\uparw}{$\uparrow$}
\newcommand{\dnarw}{$\downarrow$}
\newcommand{\ds}{DefenseSplat}
\newcommand{\dg}{DEGauss}

\definecolor{recverybad}{RGB}{210,20,20}    %
\definecolor{recbad}{RGB}{210,20,20}        %
\definecolor{recpoor}{RGB}{195,75,65}       %
\definecolor{recmid}{RGB}{170,100,95}       %
\definecolor{recgood}{RGB}{135,105,103}     %
\definecolor{recexcellent}{RGB}{105,95,94}  %

\newcommand{\defcellbold}[4]{%
  \makebox[9.4em][c]{%
    \makebox[2.2em][r]{#3}%
    \enspace
    \makebox[3.2em][r]{\brecovery{#4}}%
    \enspace
    \defbar{#1}{#2}{#3}%
  }%
}
\newcommand{\brecovery}[1]{%
  \ifdim #1 pt < 0 pt
  \textbf{\textcolor{recverybad}{(#1)}}%
  \else\ifdim #1 pt < 10 pt
  \textbf{\textcolor{recbad}{(#1)}}%
  \else\ifdim #1 pt < 25 pt
  \textbf{\textcolor{recpoor}{(#1)}}%
  \else\ifdim #1 pt < 50 pt
  \textbf{\textcolor{recmid}{(#1)}}%
  \else\ifdim #1 pt < 75 pt
  \textbf{\textcolor{recgood}{(#1)}}%
  \else
  \textbf{\textcolor{recexcellent}{(#1\%)}}%
  \fi\fi\fi\fi\fi
}
\newcommand{\recovery}[1]{%
  \ifdim #1 pt < 0 pt
  \textcolor{recverybad}{(#1)}%
  \else\ifdim #1 pt < 10 pt
  \textcolor{recbad}{(#1)}%
  \else\ifdim #1 pt < 25 pt
  \textcolor{recpoor}{(#1)}%
  \else\ifdim #1 pt < 50 pt
  \textcolor{recmid}{(#1)}%
  \else\ifdim #1 pt < 75 pt
  \textcolor{recgood}{(#1)}%
  \else
  \textcolor{recexcellent}{(#1\%)}%
  \fi\fi\fi\fi\fi
}

\definecolor{asrverylow}{HTML}{feb24c}
\definecolor{asrlow}{HTML}{fd8d3c}
\definecolor{asrmid}{HTML}{fc4e2a}
\definecolor{asrhigh}{HTML}{e31a1c}
\definecolor{asrveryhigh}{HTML}{bd0026}
\definecolor{asrmax}{HTML}{800026}

\newcommand{\asr}[1]{%
  \ifdim #1 pt < 20 pt
  {\color{asrverylow}\textbf{#1}}%
  \else\ifdim #1 pt < 40 pt
  {\color{asrlow}\textbf{#1}}%
  \else\ifdim #1 pt < 60 pt
  {\color{asrmid}\textbf{#1}}%
  \else\ifdim #1 pt < 80 pt
  {\color{asrhigh}\textbf{#1}}%
  \else\ifdim #1 pt < 95 pt
  {\color{asrveryhigh}\textbf{#1}}%
  \else
  {\color{asrmax}\textbf{#1}}%
  \fi\fi\fi\fi\fi
}

\begin{document}

\maketitle
\begin{figure}[t]
  \centering
  \includegraphics[width=\linewidth]{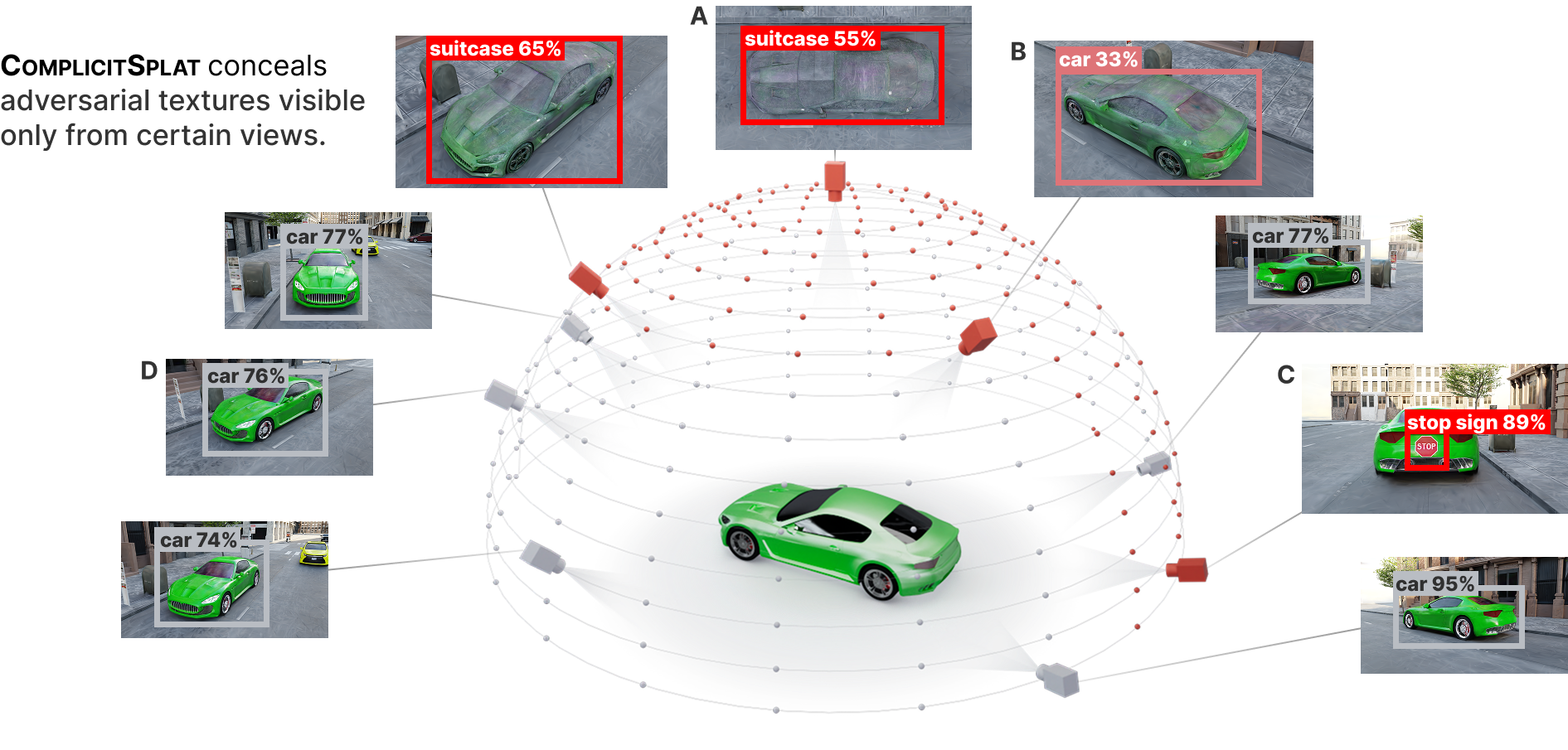}
  \caption{\textbf{\tool{}} conceals \textit{multiple} adversarial cloaked textures in 3DGS scenes using Spherical Harmonics, causing the 3DGS representation of the car to become adversarial at different view points (red dots). For example, (A) when viewed from the top, the car appears as a suitcase, (B) ``car'' detection confidence decreases, (C) and when viewed directly from behind, displays a ``stop sign.''}
  \label{fig:crown-jewel}
\end{figure}

\begin{abstract}
  As 3D Gaussian Splatting (3DGS) gains rapid adoption in safety-critical tasks for efficient novel-view synthesis from static images, how might an adversary tamper images to cause harm?
  We introduce \tool{}, the first attack to embed 2D generative edits as viewpoint-localized camouflage in 3DGS, colors and textures that change with viewing angle so that adversarial content is concealed in scene objects and visible only from attacker-chosen viewpoints, without access to downstream model weights, architecture, or the scene's training images.
  Our extensive experiments show that \tool{} generalizes across real-world captures of physical objects and synthetic indoor and outdoor scenes, successfully attacking popular single-stage, multi-stage, and transformer-based detectors while largely evading recent state-of-the-art 3DGS defenses, maintaining 97.5\% of its attack effectiveness against \textit{DEGauss} and 93.9\% against \textit{DefenseSplat}.
  To our knowledge, this is the first black-box attack on downstream object detectors via tampered 3DGS scenes, exposing a novel safety risk for applications like autonomous navigation and other mission-critical robotic systems.
  To enable reproducibility of our results, we have prepared our code for immediate release and included it in Supplementary Material.
\end{abstract}

\section{Introduction}
\label{sec:introduction}

3D Gaussian Splatting (3DGS) \cite{kerbl_3d_2023} has rapidly gained popularity in safety-critical applications due to its efficiency in novel-view synthesis from a set of static images resulting in real-time 3D rendering of complex scenes, outperforming traditional methods like Neural Radiance Fields (NeRFs) \cite{mildenhall_nerf_2020}.
The advantages of 3DGS have led to growing interest in safety-critical domains such as autonomous driving \cite{zhou_drivinggaussian_2024, li_vdg_2024,huang_s3_2024}, airborne navigation \cite{quach_gaussian_2024}, overhead (BEV) navigation \cite{lei_gaussnav_2025}, and grasping \cite{qureshi_splatsim_2024, zheng_gaussiangrasper_2024}, where rapid data generation and accurate sim2real transfer are essential.

Despite the increasing adoption of 3DGS, have the security vulnerabilities in the 3DGS scene representation been adequately considered?
Currently, 3DGS scenes are widely available for download from various sources\footnote{\url{https://poly.cam/}}\footnote{\url{https://superspl.at}}, but associated source images are often unavailable.

Given these challenges of obtaining 3DGS without source images, what harms could an attacker cause if they fine-tune a released 3DGS scene using generated training images?
We study this digital threat surface, where the attacker tampers with a distributed 3DGS asset that downstream pipelines ingest and render.
This exposes the risk of ``quiet-tampering'', \textit{i.e.}, not being able to detect that a released 3DGS scene has been adversarially fine-tuned,
resulting in \tool{}, a first-of-its-kind attack that conceals multiple adversarial appearances by exploiting the view-dependent nature of \textit{Spherical Harmonics} (SH)—a standard technique used in real-time rendering for realistic shading, enabling an attacker to subvert SH shading to embed concealed adversarial appearances into 3DGS, each visible only from specific viewing angles.
For example, Figure \ref{fig:crown-jewel} shows results of how \tool{} exploits SH shading to cause an object such as a car to appear benign from ground level and yet take on the appearance of asphalt or roadway when viewed aerially, effectively hiding from overhead surveillance systems.

\begin{figure}[t]
  \centering
  \includegraphics[width=0.85\textwidth]{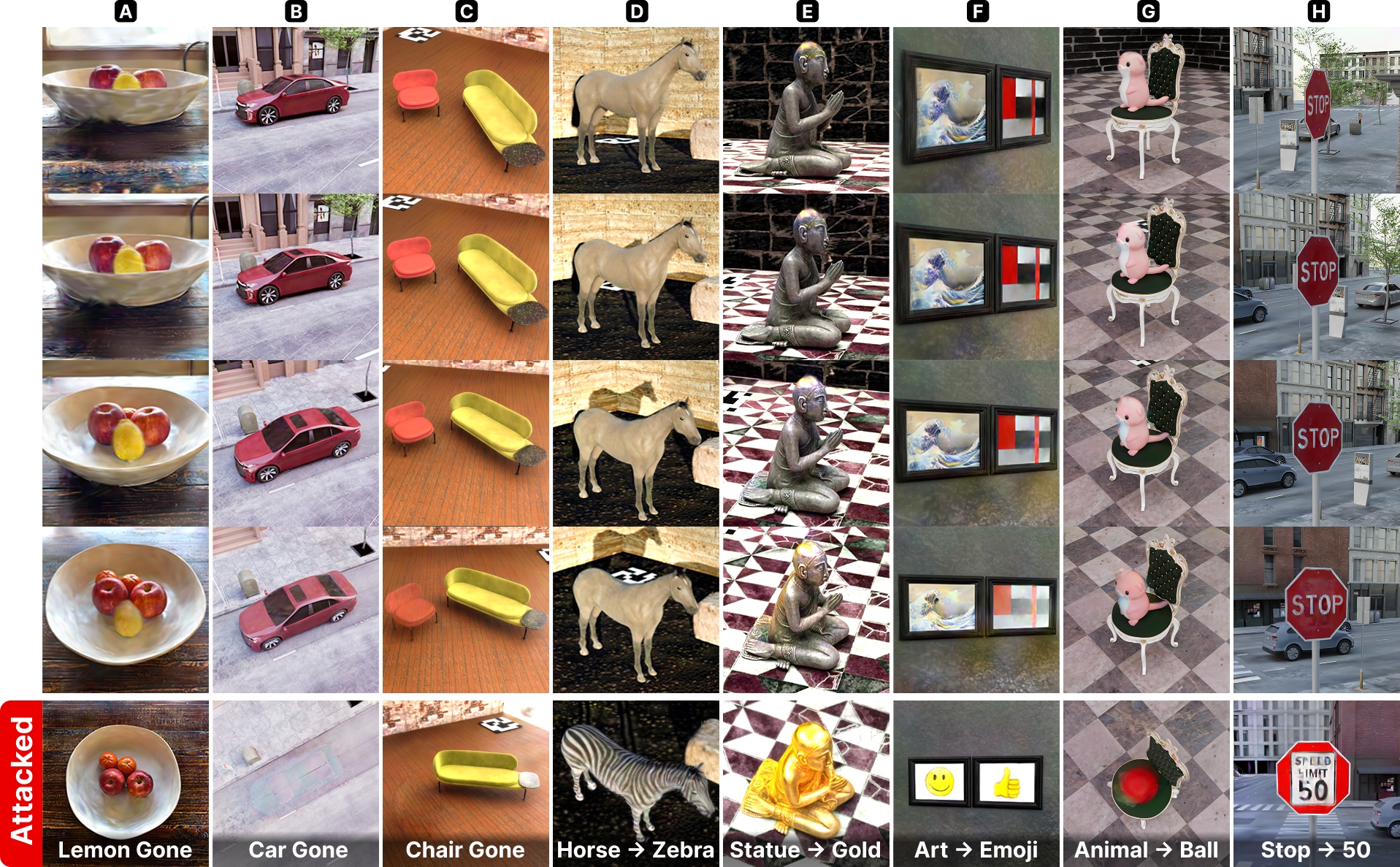}
  \caption{\tool{} generalizes across diverse objects and scenes, enabling view-dependent object-level transformations revealed only within attacker-specified angular regions. The examples demonstrate multiple transformation types, including
    \textbf{object disappearance}: lemon gone (A), car gone (B), chair gone (C);
    \textbf{texture/material alteration}: horse → zebra (D), statue → gold (E), art → emoji (F);
    \textbf{semantic substitution}: animal → ball (G); and
  \textbf{inpainting}: stop sign → ``speed limit 50'' (H).}
  \label{fig:fig_2}
  \vspace{-0.5cm}
\end{figure}

With the growing use of 3DGS in autonomous driving \cite{zhou_drivinggaussian_2024} and robotic navigation \cite{quach_gaussian_2024,lei_gaussnav_2025}, \tool{} can transform a publicly available 3DGS scene into an \textit{unknowing accomplice} by embedding adversarial, viewpoint-specific appearances that induce misclassification and missed detections in downstream object detection tasks.
The attack operates by rendering attacker-chosen viewpoints of the scene, modifying those renders with a generative model to impose a target appearance, and fine-tuning the 3DGS representation so that the altered appearance appears only within specified angular regions.
Because this process requires neither access to the original training dataset nor to downstream model architectures or weights, it demonstrates broad generalization potential across diverse object detection models, unlike prior 3DGS manipulation methods \cite{lu_poison-splat_2024,zeybey_gaussian_2024,jiang_mpam-3dgs_2025,hong_gausstrap_2025}, which assume stronger model access and training data.

Our extensive experimental results demonstrate that \tool{} generalizes across single-stage, multi-stage, and transformer-based detectors in both digitally-rendered and physically-captured 3DGS scenarios, demonstrating robust adversarial effectiveness without requiring access to internal model architectures or weights.

\begin{figure}[t]
  \centering
  \begin{minipage}[b]{0.44\textwidth}
    \centering
    \includegraphics[width=0.695\linewidth]{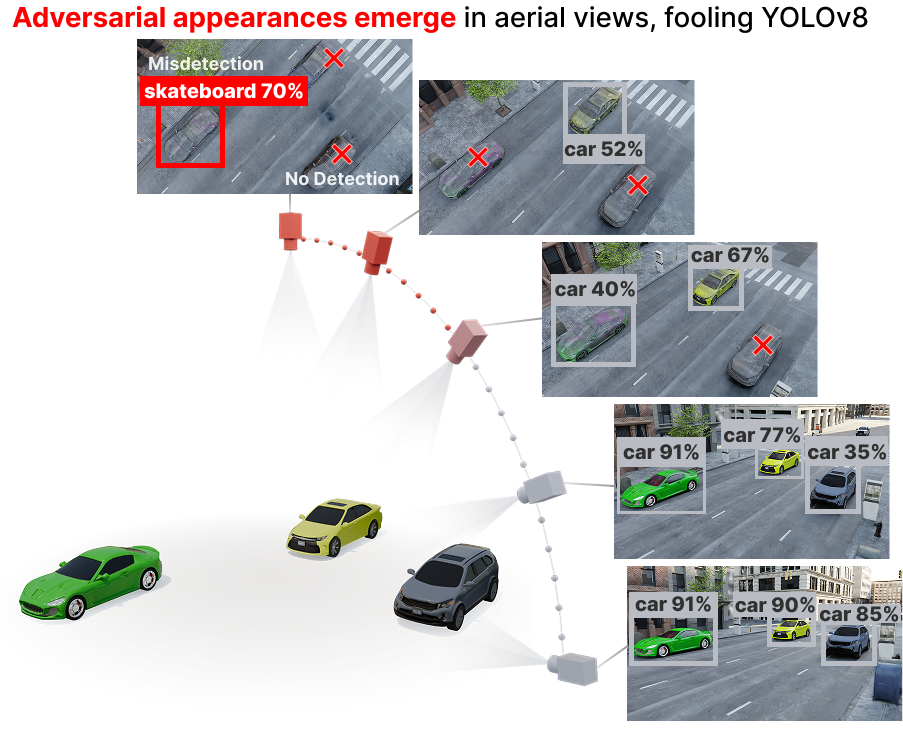}
    \caption{YOLOv8 car detections degrade and vanish as viewpoint rises to overhead.}
    \label{fig:fig_3}
  \end{minipage}\hfill
  \begin{minipage}[b]{0.52\textwidth}
    \centering
    \scriptsize
    \renewcommand{\arraystretch}{1.0}
    \captionsetup{type=table}
    \begin{tabular}{@{}l c c c c@{}}
      \textbf{Property} & \rotatebox{90}{Poison-Splat} & \rotatebox{90}{StealthAttack} & \rotatebox{90}{GSUA} & \rotatebox{90}{\textbf{Ours}} \\
      \midrule
      \tagOne{cap}   &          & \checkmark &          & \checkmark \\
      \tagTwo{cap}   &          &            & \checkmark & \checkmark \\
      Reproducible Attack & \checkmark & \checkmark   &          & \checkmark \\
      Black-Box        &          &        &          &           \checkmark \\
      \bottomrule
    \end{tabular}
    \caption{Comparison of \tool{} contributions with existing 3DGS attacks.}
    \label{tab:comparison}
    \vspace{11pt}
  \end{minipage}
\end{figure}

Recent research exploration (Table~\ref{tab:comparison}) into manipulating 3DGS remains limited --- almost no research has been open source or released publicly available code, with exception of \cite{lu_poison-splat_2024,ke2025stealthattack};
furthermore, they focus on fundamentally different goals (e.g., using extreme perturbations to drastically alter entire scene appearances or introduce severe visual artifacts)  emphasizing similarity metrics between benign and perturbed scenes rather than real-world implications for downstream tasks. \cite{zeybey_gaussian_2024,jiang_mpam-3dgs_2025,hong_gausstrap_2025}. In summary, our main contributions are:
\begin{itemize}
  \item \textbf{\tagOne{cap}}: First work showing that standard shading technique in 3DGS (spherical harmonics), can be exploited to conceal views for objects at multiple viewpoints without need to access training data as compared to other recent 3DGS attacks, such as \textit{StealthAttack}.
  \item \textbf{\tagTwo{cap}}: \tool{} transfers 2D generative edits into view-localized regions of 3DGS, evading YOLO (v3/5/8/11), Faster R-CNN, and DETR without access to their weights or architectures.
  \item \textbf{\tagThree{cap}}: Demonstrated on real-world and synthetic capture processes on indoor and outdoor 3DGS scenes, and largely evades recent defenses DefenseSplat and DEGauss, which lack public code, however, we reimplement and validate against their originally reported protective behavior before evaluating our attack.
  \item \textbf{Reproducible Attack}: Our code is ready for immediate open-source release and contains detailed attack procedures for camouflaged 3DGS fine-tuning in the Supplemental Material.
\end{itemize}

\section{Related Work}
\label{sec:related_work}
We group related research into three main areas: adversarial attacks utilizing differentiable rendering, security issues inherent in novel-view synthesis, and vulnerabilities specific to 3D Gaussian Splatting (3DGS).

\subsection{Attacks Using Differentiable Rendering}
\label{subsec:adv-attack-diff-rendering}
Adversarial attacks in the 2D space are well-established \cite{szegedy_intriguing_2014, goodfellow_explaining_2015}, and the corresponding vulnerabilities are extensively studied \cite{madry_towards_2018, carlini_evaluating_2019}.
However, such studies are not prevalent regarding 3D spaces \cite{li_survey_2024, hull_renderbender_2025}.
Attackers have used differentiable rendering methods \cite{nimier_david_mitsuba_2019, ravi_accelerating_2020, mildenhall_nerf_2020,kerbl_3d_2023} to perform adversarial gradient optimization of components in a scene, which can be used to create highly realistic scenes where perturbations are applied to geometry, texture, pose, lighting, and sensors.
This results in physically plausible objects that could be transferred to the real world \cite{zheng_physical_2024}.
Even more recently, adversarial ML researchers have used NeRF and 3DGS to extend differentiable rendering attacks to novel-view synthesis, following their rising popularity in computer vision and graphics applications \cite{irshad_neural_2024, zhu_3d_2024}.

\subsection{Security Issues in Novel-View Synthesis Methods}
\label{subsec:security-issues-3dgs}

Security vulnerabilities in novel-view synthesis methods such as 3DGS remain underexplored, but they share structural similarities with Neural Radiance Fields (NeRFs), as both rely on multi-view image supervision and known camera poses to reconstruct view-dependent scene representations.
Existing adversarial studies on NeRFs illustrate how manipulating supervision or rendered views can lead to failures in downstream perception tasks.

Prior work has demonstrated that NeRF-based systems can be exploited for facial recognition evasion via template inversion attacks, requiring no white-box access to the target recognition model \cite{shahreza_comprehensive_2023}.
NeRFail \cite{jiang_nerfail_2024} used the Iterative Gradient Signed Method (IGSM) \cite{kurakin_adversarial_2017} to generate adversarial perturbations in training images, producing NeRF reconstructions capable of misleading image classifiers.
Similarly, Wu et al.~\cite{wu_shielding_2023} applied Projected Gradient Descent (PGD) to poison training views, inducing spatial deformations in the resulting NeRF geometry.
IPA-NeRF \cite{jiang_ipa-nerf_2024} introduced a bi-level white-box optimization framework to implant viewpoint-specific backdoor appearances, revealing illusory content from selected angles while remaining hidden elsewhere, though these manipulations are largely limited to appearance insertion or disappearance.

\subsection{Adversarial Attacks on 3D Gaussian Splatting}
\label{subsec:adv-attack-3dgs}
Prior work on adversarial vulnerabilities in 3DGS remains limited.
Poison-Splat \cite{lu_poison-splat_2024} proposes a computational attack on the split/densify stage by perturbing training images to increase scene complexity, memory usage, and training time, but does not evaluate downstream perception effects.
Gaussian Splatting Under Attack (GSUA) \cite{zeybey_gaussian_2024} performs data poisoning through segmented image perturbations targeting a single classifier (CLIP ViT-B/16), without modifying the deployed 3DGS scene representation.
StealthAttack \cite{ke2025stealthattack} embeds a trigger view during training while stabilizing non-poisoned views by injecting Gaussians into low-density regions using kernel density estimation.
The archival-only GaussTrap \cite{hong_gausstrap_2025} produces hidden illusory views in trained 3DGS models, similar to IPA-NeRF backdoor mechanisms, but relies on global scene transformations.
Both evaluate success only through image similarity metrics (PSNR, SSIM, LPIPS) rather than downstream perception.
MPAM-3DGS \cite{jiang_mpam-3dgs_2025} instead attacks downstream tasks by perturbing Gaussian means, scales, rotations, spherical harmonic color, and opacity to target YOLOv5 and ResNet-101.
However, these parameter-level perturbations often introduce visible artifacts such as jagged splat boundaries and geometric misalignment, making the attack more conspicuous.

However, these works provide only limited exploration, and with the exception of \cite{lu_poison-splat_2024,ke2025stealthattack}, do not release publicly available code, hindering reproducibility and comparison between methods.
Furthermore, they emphasize extreme perturbations that drastically alter entire scene appearances or introduce visible artifacts.
Their evaluations largely rely on image-similarity metrics between benign and perturbed scenes, rather than assessing concrete impacts on downstream perception systems. Table~\ref{tab:comparison} contrasts \tool{} against existing methods. Our method:
\begin{enumerate}
  \item Embeds one or more adversarial appearances at the object level rather than overwriting the entire scene; concealments activate only within attacker-specified angular regions and maintain stealth by avoiding visible artifacts, unlike MPAM-3DGS.
  \item Demonstrates that a black-box adversarial attack can be performed by fine-tuning publicly available 3DGS scenes in both digital and real-world capture settings, without access to original training data or downstream model architectures, unlike IPA-NeRF, Poison-Splat, StealthAttack, and MPAM-3DGS.
  \item Evaluates against a broader range of object detectors than prior work and shows effectiveness across multiple architectures (YOLOv3/5/8/11, Faster R-CNN, DETR) without requiring internal model weights or architecture access.
\end{enumerate}

\begin{figure}[t]
  \centering
  \includegraphics[width=0.8\columnwidth]{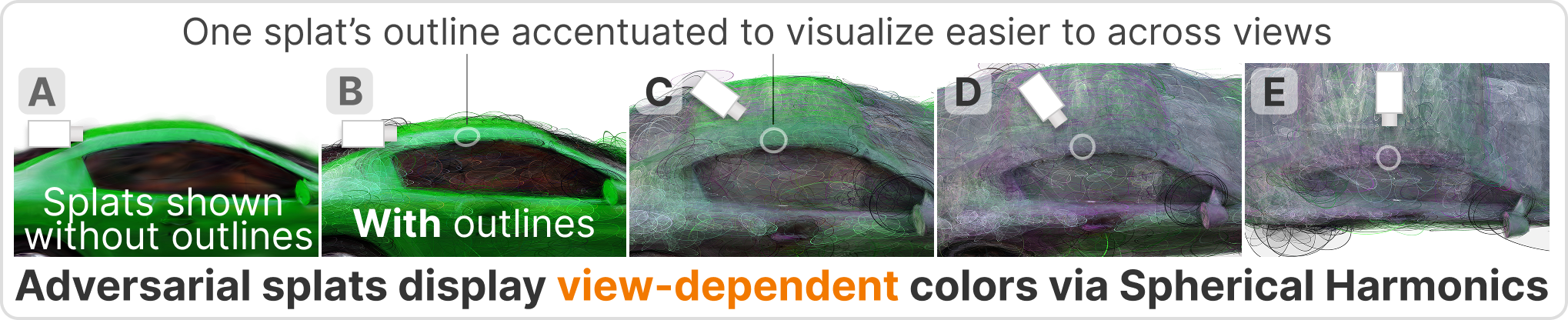}
  \caption{Adversarial Gaussian splats demonstrating view-dependent color changes enabled by spherical harmonic rendering. We highlight a single splat with a light border for easier tracking of color changes across views, revealing its transition from green to gray when rotating from a side view (frames A–B) to an overhead view (frames C–E).}
  \label{fig:fig_4}
\end{figure}

\section{Proposed Method: \tool{}}
\label{sec:intuition-threat-model}
\subsection{Main Idea: Exploit Spherical Harmonics Shading}
\label{subsec:spherical-harmonics-vulnerability}
\tool{} leverages the view-dependent properties of 3D Gaussian Splatting (3DGS) using spherical harmonics (SH) encoding to hide adversarial content within 3D scenes (Fig.~\ref{fig:fig_4}).
Spherical harmonics (SH) form an orthonormal set of basis functions commonly used to efficiently approximate diffuse lighting and shading in computer graphics.
Unlike explicit lighting models—such as Phong shading \cite{phong1998illumination}, where lighting calculations depend explicitly on known scene geometry, light positions, and viewing angles—3DGS stores precomputed SH coefficients per Gaussian splat.
This allows each splat’s appearance to change smoothly as the viewpoint shifts, without recalculating explicit light-object-camera interactions.

Typically, 3DGS is trained using SH with order $\ell=2$, yielding $(\ell+1)^2 = 9$ basis functions and 9 coefficients per color channel, totaling 27 coefficients for RGB, which is sufficient to represent most view-dependent color variation in practice \cite{green_spherical_2003}.
Reducing $\ell$ limits the expressive capacity of the appearance model; for example, $\ell=1$ uses $(\ell+1)^2 = 4$ coefficients per channel (12 total for RGB).
These coefficients parameterize a continuous directional color function that is fitted during training; SH coefficients are optimized across multiple camera views, effectively capturing realistic color variations such as reflections and intricate lighting conditions.
Notably, this process gives an adversary considerable capabilities to embed multiple adversarial appearances within a scene for some target object and viewing angles.

We exploit the view-dependent nature of SH by fine-tuning an existing 3DGS scene using generatively edited images for specific viewing angles, enabling sophisticated concealment.
For example, a car can be designed with an adversarial appearance from a top view while maintaining benign appearances from all other angles (Fig.~\ref{fig:crown-jewel}, Fig.~\ref{fig:fig_3}).
Walking 360 degrees around such a vehicle on the ground appears completely normal, as the top of the car viewed from ground level shows no indication of the hidden adversarial content.

\subsection{Threat Model}
\label{subsec:threat-model}

We formalize the black-box threat model under which \tool{} operates.

\textbf{Attacker's goals:}
The adversary aims to embed concealed adversarial content into a publicly distributed 3DGS asset that downstream pipelines ingest and render, such that the content becomes visible only from specific, attacker-chosen viewpoints.
We consider this digital threat surface only; transfer to physically printed and re-captured objects is out of scope and left to future work.
The resulting scene maintains benign appearance under normal inspection while reliably inducing misclassification or missed detections under targeted viewing conditions.

\textbf{Attacker's knowledge:}

The adversary is assumed to have only a downloaded 3DGS scene file (e.g., a \texttt{.ply}), which is inherently editable once obtained, and the ability to specify camera viewpoints for targeted angular regions and appearances via textual prompts.
They are not required to understand the 3DGS optimization procedure, nor to access the original training dataset used to construct the scene.
The adversary does not have access to the architectures or weights of any downstream models that consume 3DGS renderings.

\textbf{Attacker's capabilities:}
We model the adversary as operating under 2 bounded computational resources, reflecting realistic limits on generative editing cost and scene optimization time.
\begin{enumerate}
  \item An editing budget $B_{\text{edit}}$ that limits the total number of viewpoints whose rendered images may be modified by the generative model.
  \item An optimization budget $B_{\text{iter}}$ that limits the total number of 3DGS update iterations performed during scene refinement.
\end{enumerate}
Within these constraints, the adversary renders selected viewpoints, generates edited targets for adversarial views, and refines the 3DGS scene using the attack pipeline described in Section~\ref{sec:method}.

\subsection{Problem Formulation \& Algorithm}
\label{sec:method}
We formulate \tool{} as a black-box adversarial editing attack on a publicly available 3D Gaussian Splatting (3DGS) scene.
Let $\mathcal{S}_0$ denote the initial 3DGS scene (e.g., a \texttt{.ply} file), and let $R(\mathcal{S}, c)$ denote the differentiable renderer that produces an image from scene $\mathcal{S}$ at camera pose $c$.

\textbf{Attacker-chosen viewpoints.}
Let $C$ denote the continuous space of camera poses.
The attacker samples a finite set of viewpoints $\mathcal{C} = \{c_i\}_{i=1}^{M} \subset C$ and renders the corresponding images $x_c = R(\mathcal{S}_0, c)$ for all $c \in \mathcal{C}$.
The attacker specifies $n$ disjoint targeted angular regions $\{R_1, R_2, \dots, R_n\} \subset C$.
Each region $R_j$ may represent any attacker-defined subset of poses within $C$.
In our experiments, we instantiate $R_j$ using an angular threshold around a reference pose $c_{\text{ref}, j}$.
\begin{equation}
  \label{eq:targeted_viewpoint_regions}
  R_j = \{ c \in C : \angle(c, c_{\text{ref}, j}) \leq \delta_j \}, \quad j=1,\dots,n.
\end{equation}
In simple cases, this could correspond to a spherical cap in viewpoint space shown in Fig.~\ref{fig:camera_layout}A.
We enforce disjointness $R_j \cap R_k = \emptyset$ for $j \neq k$, since a single viewpoint cannot simultaneously correspond to multiple distinct target appearances.
For each region, we define the sampled adversarial viewpoints as $\mathcal{C}_{\text{adv}}^{(j)} = \mathcal{C} \cap R_j$ and $\mathcal{C}_{\text{adv}} = \bigcup_{j=1}^{n} \mathcal{C}_{\text{adv}}^{(j)}$.
We denote the benign viewpoint set as $\mathcal{C}_{\text{ben}} = \mathcal{C} \setminus \bigcup_{j=1}^{n} \mathcal{C}_{\text{adv}}^{(j)}$.

\textbf{Region-conditioned appearance generation.}
For each targeted region \(R_j\), the attacker specifies a prompt \(p_j\) describing the target appearance and applies a generative editor \(G\) to selected renders.
Let \(\mathcal{C}_{\text{adv}}^{(j)}\) denote the adversarial camera set for region \(R_j\).
At refresh event \(m\), the edited supervision images are
\[
  \tilde{x}_{c,m}^{(j)} = G\!\big(R(\mathcal{S}(\theta_{t_m}),c);\,p_j,\gamma_{j,m}\big),
  \qquad c\in\mathcal{C}_{\text{adv}}^{(j)},
\]
where \(t_m = mK\) is the optimization step at refresh event \(m\), \(K\) is the refresh period, and \(\gamma_{j,m}\) denotes region-conditioning controls (e.g., attention gates and re-weighting schedules).
At \(m=0\), edits are generated from the initial scene \(\mathcal{S}_0\); at later events, edits are regenerated from the current scene state \(\mathcal{S}(\theta_{t_m})\).
Thus, every \(K\) optimization steps, stale targets are replaced by refreshed \(\tilde{x}_{c,m}^{(j)}\), keeping guidance synchronized with evolving geometry.

\textbf{Fine-tuning on generated images.}
To enforce adversarial appearance within a targeted region \(R_j\) while preserving benign appearance for \(c \notin R_j\), we account for the view-dependent structure of spherical harmonic (SH) shading.
Because SH encodes appearance as a smooth, band-limited function of viewing direction, optimizing only on adversarial views causes interpolation into neighboring directions, leading to visible spillover beyond the intended region.
To localize the appearance, we jointly enforce adversarial targets in \(\mathcal{C}_{\text{adv}}^{(j)}\) and benign targets in \(\mathcal{C}_{\text{ben}}\) (Fig.~\ref{fig:camera_layout}B), especially near region boundaries, so the learned SH coefficients satisfy competing constraints. We retain the original 3DGS loss $\mathcal{L}_{3DGS} = (1 - \lambda)\mathcal{L}_1 + \lambda \mathcal{L}_{\text{D-SSIM}}$,
and optimize scene parameters \(\theta\) using supervision drawn from the region-partitioned render sets.

While the underlying 3DGS rendering formulation and parameterization remain unchanged, the training schedule is adapted to support region-partitioned refinement, including periodic supervision refresh, controlled densification windows, and weighted balancing between adversarial and benign views.
These adjustments operate within the standard 3DGS training framework and do not modify the rendering model itself. Defining $I_\theta(c)=R(\mathcal{S}(\theta),c)$ and $I_0(c)=R(\mathcal{S}_0,c)$, the attack objective becomes
\begin{equation}
  \label{eq:loss_fun}
  \min_{\theta}\;
  \alpha \sum_{c \in \mathcal{C}_{\text{adv}}}
  \mathcal{L}_{3DGS}\!\left(I_\theta(c), \tilde{x}_c \right)
  \;+\;
  \beta \sum_{c \in \mathcal{C}_{\text{ben}}}
  \mathcal{L}_{3DGS}\!\left(I_\theta(c), I_0(c) \right),
\end{equation}
where $\alpha$ and $\beta$ balance adversarial transfer against benign-view preservation.
An ablation removing the benign-view term ($\beta=0$) confirms it is the mechanism localizing the edit rather than just regularization: in-region attack strength is preserved while benign-view appearance diverges sharply outside the target region (Appendix~\ref{subsec:beta_ablation}).

\section{Evaluation}
\label{sec:evaluation}

We evaluate \tool{} from four aspects:
(i) quantitative effectiveness against downstream object detection models,
(ii) viewpoint-localized behavior under attacker-defined angular regions,
(iii) qualitative generalization across diverse scene types and transformation modalities,
and (iv) ability to evade existing 3DGS defenses.

For generative editing, we employ FLUX.2-dev\footnote{\url{https://huggingface.co/black-forest-labs/FLUX.2-dev}}, a high-capacity prompt-conditioned image editing model with a locally deployable implementation.
Its support for image-to-image editing and compatibility with multi-view processing make it suitable for controlled adversarial appearance generation within our pipeline.

\begin{figure}[t]
  \centering
  \includegraphics[width=0.8\columnwidth]{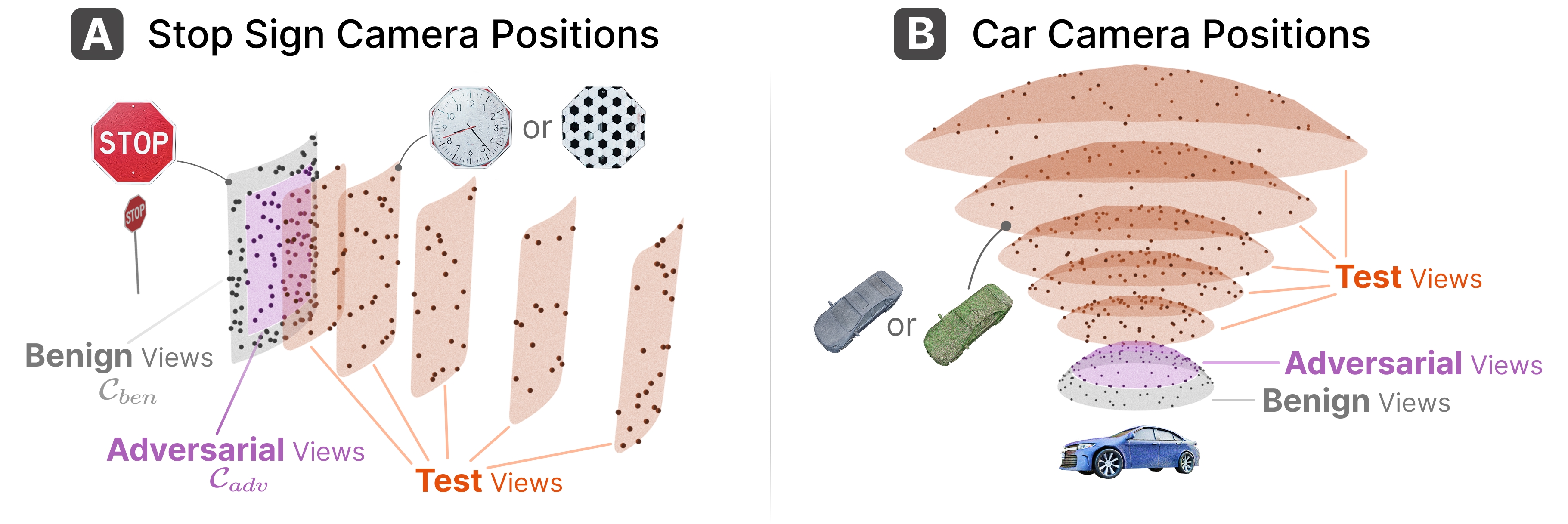}
  \caption{
    Camera layout for data collection in both scenarios.
    (A) Overhead vehicle scenario with cameras distributed across a hemisphere.
    (B) Ground-level stop sign scenario with cameras covering front-facing oriented views.
  }
  \label{fig:camera_layout}
\end{figure}

Camera sampling layouts for our primary quantitative scenarios are illustrated in Fig.~\ref{fig:camera_layout}, and representative qualitative transformations are shown in Fig.~\ref{fig:fig_2}.
Quantitative results are reported in Section~\ref{subsec:results}.

We assess the attacks against YOLO (v3, v5, v8, v11), Faster R-CNN, and DETR object detectors.
These models were selected to span a range of detection architectures:
YOLO variants represent efficient single-stage detectors \cite{cao_real-time_2023};
Faster R-CNN is a multi-stage detector with higher complexity \cite{leng_recent_2024};
and DETR \cite{carion_end--end_2020} represents transformer-based detection frameworks.

\subsection{Attacked Scenes}
\label{subsec:scene-sources}

Under the threat model described in Section~\ref{subsec:threat-model}, the adversary begins from a pretrained 3DGS scene $\mathcal{S}_0$.
In our evaluation, these scenes originate from three sources: (i) synthetic indoor and outdoor scenes rendered in Blender and converted into 3DGS representations, (ii) real-world scenes captured and reconstructed by us using consumer photogrammetry tools, and (iii) publicly available 3DGS scene files obtained from online repositories.
In all cases, the adversary applies the attack pipeline defined in Section~\ref{sec:method}: sampled viewpoints are rendered, a subset is generatively edited according to attacker-specified prompts, and the scene is refined under region-partitioned supervision within the standard 3DGS training framework.

For quantitative evaluation, we targeted two safety-critical scenarios relevant to autonomous navigation.
The first is an \textbf{overhead vehicle} scenario, where vehicles are observed from a hemispherical distribution of viewpoints.
The second is a \textbf{ground-level stop sign} scenario, where views are sampled over a front-facing arc.
\subsection{Experimental Setup}
\label{subsec:experimental-setup}
For each scene, the adversary defines an angular region $R_j \subset C$ characterized by a reference pose $c_{\text{ref}}$ (\textit{e.g.}, top-view) as described in Section~\ref{sec:method}, and then approximates each $R_j$ by sampling a finite set of adversarial viewpoints $\mathcal{C}_{\text{adv}}^{(j)} \subset R_j$ subject to the global editing budget $|\mathcal{C}_{\text{adv}}| \le B_{\text{edit}}$.

In our experiments we fix the editing budget at $B_{\text{edit}}\approx 25$ views per region (10 generative steps per image), which we found sufficient for reliable in-region transfer in preliminary runs.

Each region $R_j$ is associated with a single semantic prompt $p_j$ that defines the desired adversarial appearance across all viewpoints $c \in \mathcal{C}_{\text{adv}}^{(j)}$ in that region.
For example, in the overhead vehicle scenario, $p_j$ specifies a texture modification (e.g., converting the vehicle surface to grass or road texture), while in the stop sign scenario it specifies a semantic transformation (e.g., changing the stop sign to display a clock or soccer ball pattern).

Finally, the adversary refines the scene for up to $B_{\text{iter}} = 20{,}000$ optimization iterations.
The underlying 3DGS rendering model remains unchanged, but several training hyperparameters are adapted for adversarial refinement.
Specifically, the loss in Eq.~\ref{eq:loss_fun} is upweighted with $\alpha \in [10,25]$ while keeping $\beta = 1.0$ for benign views so adversarial targets remain competitive within $R_j$.
We also modify the densification schedule (earlier start and cutoff) to allocate Gaussian capacity for localized appearance changes while limiting artifact fitting, and perform periodic guidance refresh every $K$ iterations.
\subsection{Evaluation Metrics}
\label{subsec:evaluation-metrics}

We evaluate attack effectiveness across detectors and assess whether the refined scene accurately encodes the desired adversarial appearance using the following metrics:

\begin{itemize}
  \item \textbf{Attack Success Rate (ASR)}:
    The percentage of images where the targeted object is misclassified or not detected under adversarial conditions compared to benign conditions under a detection confidence threshold of 0.5, following Shapeshifter~\cite{chen_shapeshifter_2019}.

  \item \textbf{Average Precision (AP@0.5)}:
    The precision–recall metric at IoU threshold 0.5, following prior camouflage and adversarial object detection work \cite{suryanto_active_2023, zhou_rauca_2024}.

  \item \textbf{Perceptual Edit Fidelity (SSIM/PSNR/LPIPS)}:
    We compute metrics between the edited targets $\tilde{x}_c$ and the refined renders $I_\theta(c)$ for $c \in \mathcal{C}_{\text{adv}}$ to quantify how faithfully the optimized 3DGS scene reproduces the intended adversarial appearance in-region.
\end{itemize}

\subsection{Results}
\label{subsec:results}
We report experiments to answer the following questions:

\begin{enumerate}[label=Q\arabic*., leftmargin=7mm]
  \item \textbf{Viewpoint-Specific Camouflage}: How reliably can \tool{} induce misclassification or missed detections from attacker-chosen viewing angles while preserving benign appearance outside the targeted regions, across both synthetic and real-world indoor/outdoor 3DGS scenes (Fig.~\ref{fig:camera_layout})?
  \item \textbf{Detector Generalization}: Does \tool{} consistently evade detection across multiple object detection architectures, including lightweight single-stage (YOLO), multi-stage (Faster R-CNN), and transformer-based (DETR) models?
  \item \textbf{Perceptual Edit Fidelity}: How faithfully does the refined 3DGS scene reproduce the intended adversarial appearance in targeted regions, measured by SSIM/PSNR/LPIPS between edited targets and refined renders (and illustrated qualitatively in Fig.~\ref{fig:fig_2})?
  \item \textbf{Defense Evasion}: How well does \tool{} perform against existing state-of-the-art 3DGS defenses, \text{i.e.} DefenseSplat~\cite{qiao2026defensesplat}, DEGauss~\cite{meng2026degauss}?
\end{enumerate}

\subsection{Q1. Viewpoint-Specific Camouflage}
\label{sec:viewpoint-specific-camouflage}
The effectiveness of \tool{} in disguising targeted objects (\textit{e.g.,} vehicles as roads, stop sign as clocks) is measured by the attack success rate (ASR) on viewing angles in the test set of viewing angles.
Using the overhead based car as an example, ASR is the fraction of images where a car detected under benign conditions is not detected as a car under adversarial conditions.
Table~\ref{tab:asr_combined} reports ASR (mean and 95\% bootstrap CI, 1000 resamples over per-view detections) for the overhead car and ground stop sign.
The detectable-view count (Det.) differs across conditions because it counts only views where the object is detected under benign rendering, which varies with each detector's benign sensitivity; ASR is computed per detector over its own denominator and then pooled, so it is not dominated by any single detector.
Because nearby viewpoints are spatially correlated, these bootstrap intervals likely understate true variance and should be read as indicative rather than exact.
ASR tracks the visual dissimilarity between benign and target appearance: high-contrast disappearance edits (car$\to$grass) reliably succeed, while low-contrast material swaps (car$\to$road) preserve the object silhouette and are harder, so we expect and observe a spread across conditions.
Full per-scene results: per-detector ASR spans 11--100\% across conditions (Appendix Tables~\ref{tab:asr_stop_sign_bootstrap}--\ref{tab:asr_car_bootstrap}); the pooled estimates above summarize this spread.

\begin{table*}[t]
  \centering
  \scriptsize
  \begin{minipage}[t]{0.44\textwidth}
    \centering
    \begin{tabular}{@{}lccc@{}}
      \addlinespace[15pt]
      \textbf{Attack} & \textbf{Suc.}\,\textbf{/}\,\textbf{Det}. & \textbf{ASR} \textbf{(\%)} & \textbf{95\%} \textbf{CI} \\
      \midrule
      \CarGray{} as Grass   & 333 / 509 & 65.42 & [61.30, 69.55] \\
      \CarGray{} as Road    & 258 / 509 & 50.69 & [46.37, 55.01] \\
      \TikZStop{} as Clock  & 414 / 707 & 58.56 & [54.88, 62.09] \\
      \TikZStop{} as Soccer & 285 / 707 & 40.31 & [36.78, 43.99] \\
      \bottomrule
    \end{tabular}
    \vspace{-1pt}
    \caption{Bootstrap ASR estimates pooled over detectors. Confidence intervals are 95\% percentile bootstrap intervals over benign-detectable views.}
    \label{tab:asr_combined}
  \end{minipage}
  \hfill
  \begin{minipage}[t]{0.52\textwidth}
    \centering
    \setlength{\tabcolsep}{4pt}
    \begin{tabular}{lcccc}
      \textbf{Detector} & $\boldsymbol{N}_{\text{anchor}}$ & \textbf{ASR}$_{2D}\uparrow$ & \textbf{ASR}$_{3D}\uparrow$ & $\boldsymbol{\Delta}$\textbf{ASR} $\downarrow$\\
      \midrule
      YOLOv3     & 23 & 1.000 & 1.000 & 0.000 \\
      YOLOv5     & 28 & 1.000 & 1.000 & 0.000 \\
      YOLOv8     & 29 & 1.000 & 1.000 & 0.000 \\
      YOLOv11    & 27 & 1.000 & 1.000 & 0.000 \\
      FRCNN      & 30 & 1.000 & 0.967 & 0.033 \\
      DETR       & 28 & 1.000 & 0.929 & 0.071 \\
      \bottomrule
    \end{tabular}
    \caption{\textbf{2D Baseline ASR Comparison}. Our method preserves ASR on 2D edited views in fine-tuned 3DGS scenes.}
    \label{tab:baselineA_2d}
  \end{minipage}
\end{table*}

Overall, \tool{} achieves high pooled attack success rates, with the strongest effect on high-contrast disappearance edits (car$\to$grass, 65.4\%) and on stop-sign substitutions (clock, 58.6\%).
Low-contrast swaps that preserve object silhouette (car$\to$road, stop sign$\to$soccer) are correspondingly weaker.
Per-model results (Appendix Tables \ref{tab:asr_stop_sign_bootstrap}-\ref{tab:asr_car_bootstrap}) show expected variation across detectors; consistent with prior reports that earlier YOLO versions can be more robust in vehicle detection than later ones \cite{kilikaya_yoloperf_2023}, YOLOv3/v5 are more difficult to fool than YOLOv8/v11 on the car scenario.
\subsection{Q2. Detector Generalization}
\label{sec:detector-generalization}
We evaluate whether \tool{} consistently evades detection across diverse object detectors by measuring the change in AP@0.5 under ``road'' and ``grass'' camouflage on overhead cars and ``clock'' and ``soccer'' camouflage on stop signs (Fig.~\ref{fig:camera_layout}).
All detectors show substantial AP@0.5 reductions across scenarios. For cars, grass camouflage generally produces larger degradation than road camouflage for most detectors.
For stop signs, camouflage is even more effective: clock textures reduce AP@0.5 by up to 0.52 (YOLOv11) and 0.47 (DETR), while soccer textures cause drops of 0.34--0.43 across models.
\subsection{Q3. Perceptual Edit Fidelity}
\label{subsec:recon_fidelity}

As a proxy for edit fidelity, Table~\ref{tab:attack_metrics_traincam} splits SSIM/PSNR/LPIPS in 3 ways. The Adversarial column measures in-region fidelity: how faithfully the refined splat reproduces the intended 2D edit from adversarial views.
Benign versus Naive 3DGS compares the tampered scene's out-of-region renders against the original untampered source $S_0$. Benign metrics closely track the source, confirming fine-tuning preserves benign regions while high Adversarial scores confirm the generative edit transfers reliably into the splat.

\begin{table}[t]
  \centering
  \scriptsize
  \setlength{\tabcolsep}{3pt}
  \begin{tabular}{lcccccc|ccc}
    & \multicolumn{3}{c}{\textbf{Adversarial}} & \multicolumn{3}{c|}{\textbf{Benign}} & \multicolumn{3}{c}{\textbf{Naive 3DGS (w/o attack)}} \\
    \cmidrule(lr){2-4}\cmidrule(lr){5-7}\cmidrule(lr){8-10}
    \textbf{Scene} & \textbf{SSIM}$\uparrow$ & \textbf{PSNR}$\uparrow$ & \textbf{LPIPS}$\downarrow$
    & \textbf{SSIM}$\uparrow$ & \textbf{PSNR}$\uparrow$ & \textbf{LPIPS}$\downarrow$
    & \textbf{SSIM}$\uparrow$ & \textbf{PSNR}$\uparrow$ & \textbf{LPIPS}$\downarrow$ \\
    \midrule
    Fruit Bowl  & 0.904 & 27.416 & 0.251 & 0.898 & 26.068 & 0.081 & 0.856 & 25.501 & 0.282  \\
    Car         & 0.957 & 34.441 & 0.170 & 0.884 & 26.063 & 0.084 & 0.874 & 27.755 & 0.185 \\
    Chair       & 0.947 & 34.837 & 0.150 & 0.983 & 39.640 & 0.016 & 0.914 & 33.808 & 0.215 \\
    Horse       & 0.967 & 32.947 & 0.026 & 0.877 & 29.748 & 0.066 & 0.869 & 27.509 & 0.155 \\
    Statue      & 0.925 & 26.939 & 0.056 & 0.944 & 33.204 & 0.039 & 0.934 & 35.861 & 0.023 \\
    Art         & 0.905 & 21.414 & 0.120 & 0.926 & 25.423 & 0.102 & 0.954 & 39.221 & 0.213 \\
    Squirrel    & 0.934 & 27.608 & 0.083 & 0.951 & 32.137 & 0.063 & 0.908 & 21.253 & 0.177 \\
    Stop Sign   & 0.983 & 36.911 & 0.018 & 0.959 & 31.511 & 0.033 & 0.972 & 37.482 & 0.070 \\
  \end{tabular}
  \caption{Attack-view reconstruction metrics split by adversarial, benign, and naive 3DGS (w/o attack) views. Higher SSIM/PSNR and lower LPIPS are better.}
  \label{tab:attack_metrics_traincam}
\end{table}
To further contextualize these reconstruction results, we also compare against StealthAttack~\cite{ke2025stealthattack}, by reproducing their poisoning setting within our pipeline by fine-tuning scenes directly on the poisoned image, rather than applying our region-conditioned black-box attack setup.

\begin{figure}[t]
  \centering
  \begin{minipage}[c]{0.48\textwidth}
    \centering
    \includegraphics[width=\linewidth]{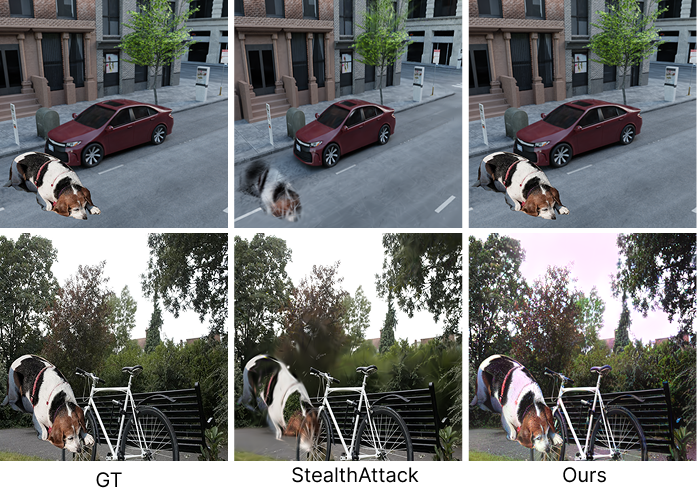}
  \end{minipage}\hfill
  \begin{minipage}[c]{0.48\textwidth}
    \centering
    \scriptsize
    \setlength{\tabcolsep}{4pt}
    \begin{tabular}{llccc}
      \toprule
      \textbf{Scene} & \textbf{Method} & \textbf{SSIM}$\uparrow$ & \textbf{PSNR}$\uparrow$ & \textbf{LPIPS}$\downarrow$ \\
      \midrule
      \multirow{2}{*}{Bicycle}
      & StealthAttack & 0.674 & 20.670 & 0.323 \\
      & \textbf{Ours} & \textbf{0.696} & 19.124 & \textbf{0.191} \\
      \midrule
      \multirow{2}{*}{Car}
      & StealthAttack & 0.847 & 26.783 & 0.300 \\
      & \textbf{Ours} & \textbf{0.999} & \textbf{50.871} & \textbf{0.003} \\
      \bottomrule
    \end{tabular}
  \end{minipage}
  \caption{Qualitative comparison on poisoned scenes. Compared with StealthAttack (center), our reproduction (right) yields consistently clearer objects with fewer artifacts, on poisoned car and mip-NeRF-360 bicycle scenes~\cite{ke2025stealthattack}.}
  \label{fig:stealthattack_comparison}
  \vspace{-0.5cm}
\end{figure}

\vspace{-0.4cm}
\subsection{Defense Evasion}
Defense study on 3DGS is very limited, and only 2 defenses are known: DefenseSplat~\cite{qiao2026defensesplat}, a white-box defense applying discrete wavelet transform (DWT) low-pass filtering to counter the PoisonSplat~\cite{lu_poison-splat_2024} attack on training images, and DEGauss~\cite{meng2026degauss}, which defends against malicious multi-view editing by optimizing 3D perturbations that disrupt future editing guidance.
Neither released code, so we reimplement both from their papers and validate each reimplementation against the metrics each paper reports.
Our DefenseSplat reproduction matches the published filtering effect, recovering the reported Gaussian count reduction and PSNR/SSIM/LPIPS gains on Tanks\&Temples and Mip-NeRF~360 (Appendix Table~\ref{tab:defensesplat-reproduction-fidelity}).
Our DEGauss reproduction matches the reported preservation (PSNR within $0.07$) and is no stronger than the original on editing-disruption metrics, so evasion is not an artifact of a weakened defense (Appendix Table~\ref{tab:degauss_reproduction_fidelity_three_scenes}).

For DefenseSplat, we render an attacked scene from the Fig.~\ref{fig:camera_layout} camera positions, apply DWT low-pass filtering to the renders, and train a new 3DGS on the defended images; we then run the detector suite per our threat model (\ref{subsec:threat-model}) and compare to baseline and attacked results (Table~\ref{tab:defensesplat_with_benign}).
DEGauss operates directly on \texttt{.ply} data as a protective step; we protect benign scenes using their 2000-step setup, attack the protected scenes with \tool{}, and evaluate as above.

\tool{} largely evades both defenses, which recover only a small fraction of lost detection (Table~\ref{tab:defensesplat_with_benign}).
Overall, DefenseSplat restores 6.1\% of the gap and DEGauss only 2.5\%.
For disappearance attacks (Car, Chair) neither defense meaningfully recovers, and in two cases (Car$\to$Grass$\vert$Road under \dg) it lowers detection below the attacked baseline.
For visually dissimilar targeted attacks (Stop sign, Horse) defenses fare better, with \dg\ recovering up to 52.7\% on Stop sign$\to$Sports ball, yet defended rates (42.6--57.1\%) stay well below the 70.1--82.5\% benign baseline, unacceptable in safety-critical settings~\cite{eykholt2018robust,cao2019adversarial}.

\begin{table}[t]
  \centering
  \scriptsize

  \newcommand{\defbar}[3]{%
    \tikz[baseline=-0.55ex]{
      \pgfmathsetmacro{\pos}{(#3 - #1) / (#2 - #1)}
      \pgfmathsetmacro{\clippos}{min(max(\pos,0),1)}
      \draw[black!20, line width=0.65pt] (0,0) -- (2.6em,0);
      \draw[black!45, line width=0.65pt] (2.6em,-0.38ex) -- (2.6em,0.38ex);
      \fill[black!65] ({2.6em * \clippos},0) circle[radius=0.38ex];
    }%
  }

  \newcommand{\defcell}[4]{%
    \makebox[9.4em][c]{%
      \makebox[2.2em][r]{#3}%
      \enspace
      \makebox[3.2em][r]{\recovery{#4}}%
      \enspace
      \defbar{#1}{#2}{#3}%
    }%
  }

  \begin{tabular}{llcccc}
    &               & \multicolumn{4}{c}{\textbf{Detection rates (\%)}}                             \\
    \cmidrule(lr){3-6}
    \textbf{Scene} & \textbf{Attack target} & \textbf{Benign} \uparw & \textbf{Attacked} \dnarw
    & \textbf{\ds} \uparw\ \textbf{(Recovered)}
    & \textbf{\dg} \uparw\ \textbf{(Recovered)}                                         \\
    \midrule
    Car       & Grass         & 52.8          & 24.8
    & \defcell{24.8}{52.8}{13.2}{-41.4}
    & \defcell{24.8}{52.8}{0.1}{-88.2}                                                    \\
    Car       & Road          & 52.8          & 28.7
    & \defcell{28.7}{52.8}{31.2}{10.4}
    & \defcell{28.7}{52.8}{3.6}{-104.1}                                                    \\
    Stop Sign & Clock         & 82.5          & 26.3
    & \defcell{26.3}{82.5}{32.7}{11.4}
    & \defcell{26.3}{82.5}{42.6}{29.0}                                                     \\
    Stop Sign & Sports ball   & 82.5          & 28.8
    & \defcell{28.8}{82.5}{46.7}{33.3}
    & \defcell{28.8}{82.5}{57.1}{52.7}                                                     \\
    Chair     & Flooring      & 82.4          & 10.5
    & \defcell{10.5}{82.4}{12.8}{3.2}
    & \defcell{10.5}{82.4}{13.6}{4.3}                                                      \\
    Horse     & Zebra         & 64.4          & 34.6
    & \defcell{34.6}{64.4}{32.6}{-6.7}
    & \defcell{34.6}{64.4}{42.9}{27.9}                                                     \\
    \midrule
    Overall   & --            & 70.1          & 25.6
    & \defcellbold{25.6}{70.1}{28.3}{6.1}
    & \defcellbold{25.6}{70.1}{26.7}{2.5}                                                      \\
    \bottomrule
  \end{tabular}
  \caption{
    Detection rates (\%) before and after defense, where (Recovered) is
    fraction of attack-lost detection restored by the defense: 100\% = fully restored to benign,
    0\% = no improvement over attacked, negative = defense worsens detection further.
    \tool{} evades both \ds\ and \dg, with defenses recovering on average only 6.1\% (\ds) and 2.5\% (\dg).
    For disappearance attacks (Car, Chair), defenses recover at most 10.4\% and in two cases negative recovery indicates the defense actively lowers detection below the attacked baseline.
    For targeted attacks across visually dissimilar categories (Stop Sign, Horse), recovery reaches up to 52.7\%, yet defended detection rates remain well below benign levels.
  }
  \label{tab:defensesplat_with_benign}
  \vspace{-0.5cm}
\end{table}
\subsection{Ablations}
\label{subsec:ablation}

In our experimental setup, we explored the impact of two factors on the effectiveness of our adversarial camouflage attacks, leading us to conduct ablation studies on the following:
\begin{itemize}
  \item \textbf{Number of Spherical Harmonics Coefficients} Spherical harmonics (SH) coefficients determine the complexity of the camouflage appearance.
    Higher SH orders allow for more detailed estimation of object colors during training, potentially allowing better capture of camouflage patterns and increasing the effectiveness of the attack.
  \item \textbf{Camera Distances from Target Object} The distance of the camera from the target object influences the visibility and effectiveness of the camouflage.
\end{itemize}
\vspace{-0.5cm}
\subsubsection{SH Order Ablation}
\label{subsubsec:sh_order_ablation}
To ablate the number of spherical harmonics coefficients, we varied the SH order ($\ell=0,1,2$) used in 3DGS training and evaluated AP@0.5 for the \CarGray{} car under ``grass'' camouflage, using the same camera poses as the main experiments (Fig.~\ref{fig:camera_layout}-left).
Table~\ref{tab:sh_ablation} reports benign and attacked in-region AP@0.5 across orders.
Benign AP is unchanged across $\ell$, so the attacked-column drops reflect the camouflage rather than scene fidelity; since lowering $\ell$ does not consistently weaken the attack (e.g., YOLOv8 holds at 0.02--0.03), a defender cannot mitigate it by retraining at lower SH order.
While in-region strength is largely order-independent, \textit{view-dependent SH} capacity is what enables concealment: it is the benign-view term (Appendix~\ref{subsec:beta_ablation}) that provides capability to keep benign views clean.

\begin{table}[ht]
  \centering
  \scriptsize
  \begin{tabular}{lcccccc}
    \textbf{SH Order} & \textbf{YOLOv3} & \textbf{YOLOv5} & \textbf{YOLOv8} & \textbf{YOLOv11} & \textbf{FRCNN} & \textbf{DETR} \\
    \midrule
    $\ell=2$ &
    0.79 / 0.49 &
    0.74 / 0.27 &
    0.31 / 0.02 &
    0.56 / 0.28 &
    0.40 / 0.05 &
    0.35 / 0.11 \\

    $\ell=1$ &
    0.79 / 0.50 &
    0.74 / 0.24 &
    0.31 / 0.02 &
    0.56 / 0.32 &
    0.40 / 0.04 &
    0.35 / 0.13 \\

    $\ell=0$ &
    0.79 / 0.48 &
    0.74 / 0.29 &
    0.31 / 0.03 &
    0.56 / 0.30 &
    0.40 / 0.05 &
    0.35 / 0.20 \\
    \bottomrule
  \end{tabular}
  \caption{AP@0.5 for \CarGray{} car with SH ablations, reported benign / attacked for each detector.}
  \label{tab:sh_ablation}
\end{table}
\vspace{-0.5cm}
\subsubsection{Camera Distance Ablation.}
\label{subsubsec:camera_distance_ablation}
To study the effect of camera distance on adversarial camouflage, we evaluate each of the five partial hemispheres shown in Fig.~\ref{fig:camera_layout}-left independently.

For the \CarGray{} car under ``grass'' camouflage, we measure AP and AR at IoU 0.5 and average across detectors, reporting by mean camera altitude (m) above the car (Table~\ref{tab:alt_ablation}).
The attack produces substantial degradation at every altitude (\(\Delta\)AP between $-0.19$ and $-0.30$), demonstrating the camouflage is effective across the full range of overhead viewpoints rather than only at specific elevations.
\begin{table}[ht]
  \centering
  \scriptsize

  \begin{tabular}{cccc}
    \toprule
    \textbf{Altitude (m)} & \textbf{Benign AP/AR} & \textbf{Adv AP/AR} & $\boldsymbol{\Delta}$\textbf{AP / }$\boldsymbol{\Delta}$\textbf{AR} \\
    \midrule
    12 & 0.218 / 0.214 & 0.033 / 0.031 & {\color[HTML]{FC4E2A}\textbf{-0.185 / -0.184}} \\
    16 & 0.294 / 0.289 & 0.025 / 0.021 & {\color[HTML]{E31A1C}\textbf{-0.269 / -0.268}} \\
    20 & 0.442 / 0.440 & 0.199 / 0.196 & {\color[HTML]{BD0026}\textbf{-0.243 / -0.244}} \\
    24 & 0.349 / 0.349 & 0.050 / 0.048 & {\color[HTML]{800026}\textbf{-0.300 / -0.301}} \\
    30 & 0.366 / 0.363 & 0.158 / 0.156 & {\color[HTML]{800026}\textbf{-0.207 / -0.208}} \\
    \bottomrule
  \end{tabular}
  \caption{Camera distance ablation and effect on AP/AR.}
  \label{tab:alt_ablation}
\end{table}
\vspace{-1cm}
\subsection{Computational Efficiency}
\label{subsec:computational_efficiency}
On a single NVIDIA H200 GPU, generating adversarial appearances with \tool{} requires $\sim$25 edited views per region with 10 FLUX.2-dev steps each ($\sim$1\,s/step), or $\sim$250\,s per guidance refresh.
Across 4 refresh cycles, this adds $\sim$1000\,s of editing overhead.
Fine-tuning requires $\sim$20\,minutes for 20K optimization iterations, yielding a total attack generation time of $\sim$36\,minutes per scene.

For comparison, we reproduced StealthAttack~\cite{ke2025stealthattack}: the reported 30K-iteration optimization stage takes $\sim$25\,minutes, but in reproducing their pipeline we found an additional voxelization stage, not accounted for in their reported cost, requiring $\sim$2.5\,hours per mip-NeRF-360 scene.
Because \tool{} operates directly on the splat representation, it avoids this costly preprocessing stage and achieves substantially lower end-to-end attack generation time.

\vspace{-0.5cm}
\section{Conclusions and Future Directions for Our Community}
\label{sec:conclusion}
We introduced \tool{}, the first black-box attack that embeds 2D generative edits as viewpoint-localized 3DGS camouflage.
By exploiting the view-dependent properties of SH, the attack embeds adversarial appearances that activate at attacker-selected viewpoints while preserving benign appearance elsewhere.
Experiments show that these edits transfer reliably into the optimized 3DGS representation, achieving high ASR across multiple detectors without requiring access to training data or model weights, revealing a new attack surface for distributed neural-scene assets increasingly used to generate training data and test perception in robotics, autonomous navigation, and simulation.
We hope this work motivates further research on defenses and evaluation protocols for securing emerging reconstructive scene representations, as well as on vulnerabilities in downstream tasks such as segmentation, tracking, and closed-loop planning and control.
\bibliography{egbib}
\newpage
\section{Appendix}
\setcounter{page}{1}

\begin{figure}[H]
  \centering
  \includegraphics[width=0.85\textwidth]{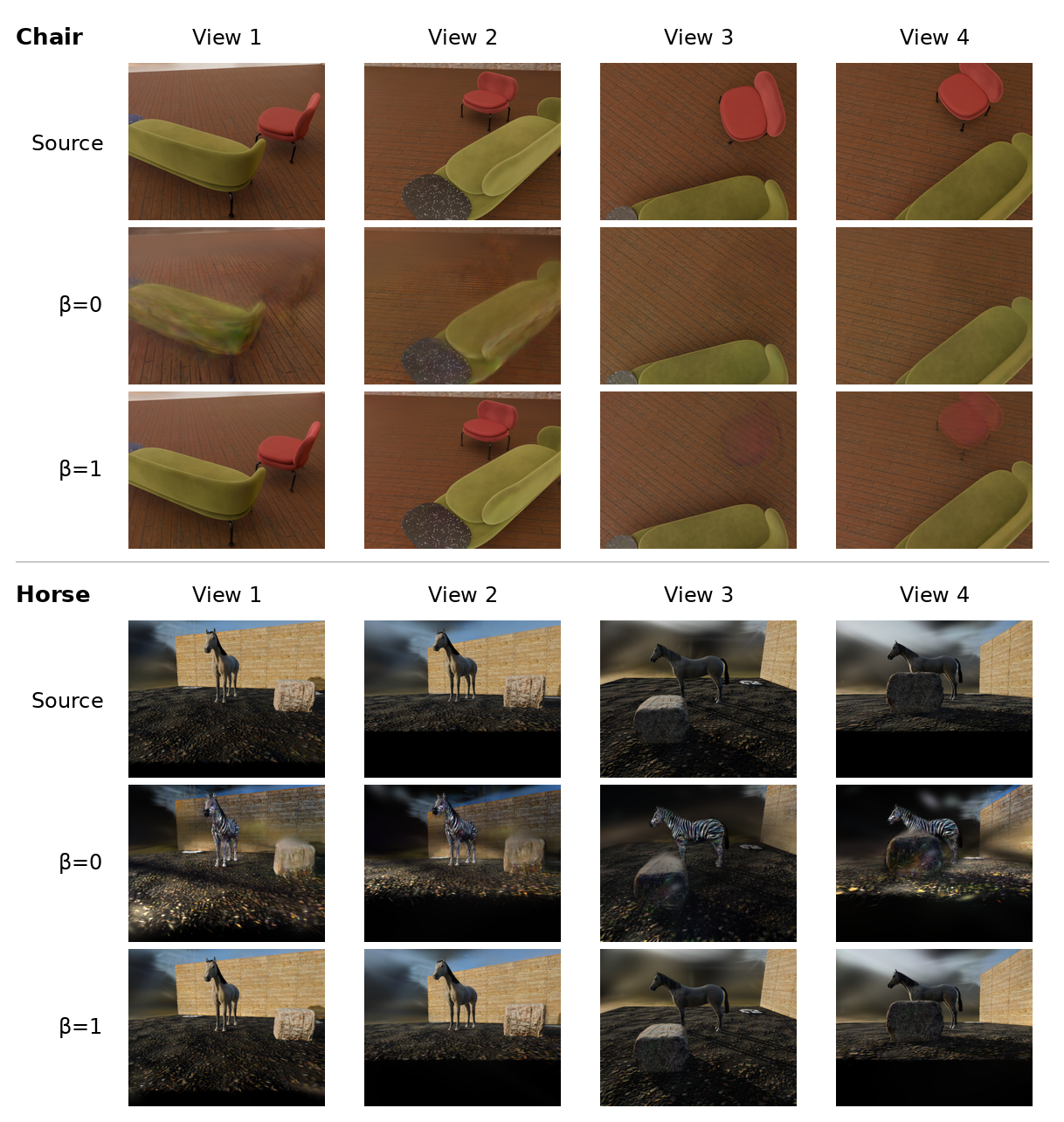}
  \caption{Effect of the benign-view term on spillover. Boundary-region renders for $\beta=1$ (full objective, top) and $\beta=0$ (adversarial-only, bottom). Without the benign term, the adversarial appearance bleeds across the region boundary into benign views (e.g., Chair/Horse on $\beta=0$ row show spillover) when boundary is constrained to overhead views while $\beta=1$ demonstrates benign appearance is maintained, and in-region edit constrained to overhead views, i.e., as in Fig.~\ref{fig:fig_2}.}
  \label{fig:beta_boundary}
\end{figure}

\subsection{Benign-View Term Ablation}
\label{subsec:beta_ablation}
The benign supervision term in Eq.~\ref{eq:loss_fun} localizes the edit in view space.
To test whether this term is what localizes the edit rather than just regularizing the fit, we compare the full objective ($\beta=1$) against an adversarial-only ablation ($\beta=0$) on the Chair and Horse scenes, and measure appearance divergence from the untampered source via LPIPS/SSIM, stratified by angular distance from the attack region (boundary $\leq 5$ views, far otherwise).
The in-region appearance divergence is comparable across conditions (e.g., horse attack-view LPIPS $0.156$ vs.\ $0.134$), so the two settings impose a similar in-region edit; the ablation isolates the effect on benign views.

Removing the term causes large view drift, strongest near the boundary
(horse boundary LPIPS $0.363 \to 0.065$, SSIM $0.449 \to 0.847$ when the term is restored; green couch $0.178 \to 0.019$), demonstrating that $\beta$ confines the adversarial appearance to the targeted region rather than letting it spill into neighbouring views.
This complements the SH-order ablation (Table~\ref{tab:sh_ablation}): whereas in-region attack strength is largely order-independent, view-dependent SH capacity is what enables \emph{concealment}, and the benign term is the mechanism that recruits it correctly.
Qualitative boundary sweeps (Fig.~\ref{fig:beta_boundary}) show the spillover directly.

\label{sec:appendix}
\newcommand{\asrci}[3]{\asr{#1} {\scriptsize [#2, #3]}}

\begin{table}[H]
  \centering
  \small
  \begin{tabular}{llrrrr}
    \textbf{Scene} & \textbf{Condition} & \multicolumn{2}{c}{\textbf{LPIPS} $\downarrow$} & \multicolumn{2}{c}{\textbf{SSIM} $\uparrow$} \\
    \cmidrule(lr){3-4}\cmidrule(lr){5-6}
    & & \textbf{Boundary} & \textbf{Far} & \textbf{Boundary} & \textbf{Far} \\
    \midrule
    Green couch & $\beta=0$ & 0.178 & 0.334 & 0.920 & 0.781 \\
    Green couch & $\beta=1$ & 0.019 & 0.043 & 0.984 & 0.965 \\
    Horse & $\beta=0$ & 0.363 & 0.267 & 0.449 & 0.655 \\
    Horse & $\beta=1$ & 0.065 & 0.078 & 0.847 & 0.893 \\
    \bottomrule
  \end{tabular}
  \caption{Ablation Study on $\beta$ supervision term on restricting adversarial editing boundaries}
  \label{tab:beta_ablation}
\end{table}

\subsection{Per-model attack success rates}
Tables~\ref{tab:asr_stop_sign_bootstrap}--\ref{tab:asr_car_bootstrap} give the per-detector ASR with 95\% percentile bootstrap confidence intervals that underlie the detector-pooled estimates in main-body Table~\ref{tab:asr_combined}.

\begin{table}[H]
  \centering
  \scriptsize
  \begin{tabular}{@{}lcccc@{}}
    \toprule
    & \multicolumn{2}{c}{\TikZStop{} as Clock}
    & \multicolumn{2}{c}{\TikZStop{} as Soccer} \\
    \cmidrule(lr){2-3}\cmidrule(lr){4-5}
    \textbf{Model} & \textbf{Suc.\,/\,Det.} & \textbf{ASR [95\% CI]}
    & \textbf{Suc.\,/\,Det.} & \textbf{ASR [95\% CI]} \\
    \midrule
    YOLOv3  & 59 / 123 & \asrci{47.97}{39.02}{56.91}
    & 24 / 123 & \asrci{19.51}{13.01}{26.83} \\

    YOLOv5  & 58 / 120 & \asrci{48.33}{39.17}{57.50}
    & 48 / 120 & \asrci{40.00}{30.83}{48.33} \\

    YOLOv8  & 68 / 117 & \asrci{58.12}{48.72}{66.67}
    & 56 / 117 & \asrci{47.86}{38.46}{56.41} \\

    YOLOv11 & 72 / 114 & \asrci{63.16}{53.51}{71.93}
    & 59 / 114 & \asrci{51.75}{42.98}{60.53} \\

    FRCNN   & 88 / 105 & \asrci{83.81}{76.19}{90.48}
    & 49 / 105 & \asrci{46.67}{37.14}{56.19} \\

    DETR    & 69 / 128 & \asrci{53.91}{45.31}{62.50}
    & 49 / 128 & \asrci{38.28}{30.47}{46.88} \\
    \bottomrule
  \end{tabular}
  \caption{Stop-sign ASR by model with 95\% percentile bootstrap confidence intervals over benign-detectable views.}
  \label{tab:asr_stop_sign_bootstrap}
\end{table}

\begin{table}[H]
  \centering
  \scriptsize
  \begin{tabular}{@{}lcccc@{}}
    \toprule
    & \multicolumn{2}{c}{\CarGray{} as Grass}
    & \multicolumn{2}{c}{\CarGray{} as Road} \\
    \cmidrule(lr){2-3}\cmidrule(lr){4-5}
    \textbf{Model} & \textbf{Suc.\,/\,Det.} & \textbf{ASR [95\% CI]}
    & \textbf{Suc.\,/\,Det.} & \textbf{ASR [95\% CI]} \\
    \midrule
    YOLOv3  & 54 / 126 & \asrci{42.86}{34.13}{51.59}
    & 14 / 126 & \asrci{11.11}{5.56}{16.67} \\

    YOLOv5  & 77 / 119 & \asrci{64.71}{56.30}{73.11}
    & 37 / 119 & \asrci{31.09}{23.53}{39.50} \\

    YOLOv8  & 49 / 50  & \asrci{98.00}{94.00}{100.00}
    & 50 / 50  & \asrci{100.00}{100.00}{100.00} \\

    YOLOv11 & 53 / 92  & \asrci{57.61}{47.83}{67.39}
    & 75 / 92  & \asrci{81.52}{72.83}{89.13} \\

    FRCNN   & 57 / 65  & \asrci{87.69}{78.46}{95.38}
    & 58 / 65  & \asrci{89.23}{81.54}{95.38} \\

    DETR    & 43 / 57  & \asrci{75.44}{63.16}{85.96}
    & 24 / 57  & \asrci{42.11}{29.82}{54.39} \\
    \bottomrule
  \end{tabular}
  \caption{Car ASR by model with 95\% percentile bootstrap confidence intervals over benign-detectable views.}
  \label{tab:asr_car_bootstrap}
\end{table}

\clearpage
\begingroup
\raggedbottom
\subsection{Per-detector AP/AR breakdowns}
Tables~\ref{tab:car-ap-ar}--\ref{tab:bluecar-ap-ar} report per-detector AP@0.5/AR@0.5 for each scene, supporting the detector-generalization analysis of Q2 (\S\ref{sec:detector-generalization}): the car table contrasts road and grass camouflage, the stop-sign table contrasts clock and soccer textures, and the blue-car table reports the real-world overhead scenario.

\begin{table}[H]
  \centering
  \tiny
  \begin{tabular}{lllcccr}
    \toprule
    \textbf{Model} & \textbf{Object} & \textbf{Appearance} & \textbf{AP@0.5} & $\boldsymbol{\Delta}$\textbf{AP} & \textbf{AR@0.5} & $\boldsymbol{\Delta}$\textbf{AR} \\
    \midrule
    \textbf{YOLOv3}
    & \CarBlue & Benign     & 0.792  &                & 0.797  &                \\
    &          & Road       & 0.733  & \textbf{-0.059} & 0.738  & \textbf{-0.059} \\
    &          & Grass      & 0.485  & \textbf{-0.307} & 0.481  & \textbf{-0.316} \\
    & \CarRed  & Benign     & 0.743  &                & 0.747  &                \\
    &          & Road       & 0.673  & \textbf{-0.070} & 0.677  & \textbf{-0.070} \\
    &          & Grass      & 0.485  & \textbf{-0.258} & 0.481  & \textbf{-0.266} \\
    & \CarGray & Benign     & 0.703  &                & 0.708  &                \\
    &          & Road       & 0.515  & \textbf{-0.188} & 0.519  & \textbf{-0.189} \\
    &          & Grass      & 0.485  & \textbf{-0.218} & 0.481  & \textbf{-0.227} \\
    \midrule
    \textbf{YOLOv5}
    & \CarBlue & Benign     & 0.743  &                & 0.744  &                \\
    &          & Road       & 0.545  & \textbf{-0.198} & 0.547  & \textbf{-0.197} \\
    &          & Grass      & 0.267  & \textbf{-0.476} & 0.266  & \textbf{-0.478} \\
    & \CarRed  & Benign     & 0.733  &                & 0.736  &                \\
    &          & Road       & 0.614  & \textbf{-0.119} & 0.613  & \textbf{-0.123} \\
    &          & Grass      & 0.307  & \textbf{-0.426} & 0.303  & \textbf{-0.433} \\
    & \CarGray & Benign     & 0.703  &                & 0.708  &                \\
    &          & Road       & 0.465  & \textbf{-0.238} & 0.468  & \textbf{-0.240} \\
    &          & Grass      & 0.257  & \textbf{-0.446} & 0.256  & \textbf{-0.452} \\
    \midrule
    \textbf{YOLOv8}
    & \CarBlue & Benign     & 0.317  &                & 0.316  &                \\
    &          & Road       & 0.010  & \textbf{-0.307} & 0.007  & \textbf{-0.309} \\
    &          & Grass      & 0.020  & \textbf{-0.297} & 0.013  & \textbf{-0.303} \\
    & \CarRed  & Benign     & 0.465  &                & 0.469  &                \\
    &          & Road       & 0.109  & \textbf{-0.356} & 0.103  & \textbf{-0.366} \\
    &          & Grass      & 0.188  & \textbf{-0.277} & 0.190  & \textbf{-0.279} \\
    & \CarGray & Benign     & 0.307  &                & 0.302  &                \\
    &          & Road       & 0.010  & \textbf{-0.297} & 0.006  & \textbf{-0.296} \\
    &          & Grass      & 0.079  & \textbf{-0.228} & 0.079  & \textbf{-0.223} \\
    \midrule
    \textbf{YOLOv11}
    & \CarBlue & Benign     & 0.564  &                & 0.568  &                \\
    &          & Road       & 0.109  & \textbf{-0.455} & 0.106  & \textbf{-0.462} \\
    &          & Grass      & 0.277  & \textbf{-0.287} & 0.275  & \textbf{-0.293} \\
    & \CarRed  & Benign     & 0.663  &                & 0.663  &                \\
    &          & Road       & 0.287  & \textbf{-0.376} & 0.283  & \textbf{-0.380} \\
    &          & Grass      & 0.455  & \textbf{-0.208} & 0.450  & \textbf{-0.213} \\
    & \CarGray & Benign     & 0.465  &                & 0.463  &                \\
    &          & Road       & 0.069  & \textbf{-0.396} & 0.068  & \textbf{-0.395} \\
    &          & Grass      & 0.327  & \textbf{-0.138} & 0.323  & \textbf{-0.140} \\
    \midrule
    \textbf{FRCNN}
    & \CarBlue & Benign     & 0.396  &                & 0.396  &                \\
    &          & Road       & 0.069  & \textbf{-0.327} & 0.067  & \textbf{-0.329} \\
    &          & Grass      & 0.050  & \textbf{-0.346} & 0.049  & \textbf{-0.347} \\
    & \CarRed  & Benign     & 0.406  &                & 0.409  &                \\
    &          & Road       & 0.079  & \textbf{-0.327} & 0.073  & \textbf{-0.336} \\
    &          & Grass      & 0.178  & \textbf{-0.228} & 0.171  & \textbf{-0.238} \\
    & \CarGray & Benign     & 0.317  &                & 0.317  &                \\
    &          & Road       & 0.109  & \textbf{-0.208} & 0.104  & \textbf{-0.213} \\
    &          & Grass      & 0.089  & \textbf{-0.228} & 0.085  & \textbf{-0.232} \\
    \midrule
    \textbf{DETR}
    & \CarBlue & Benign     & 0.347  &                & 0.348  &                \\
    &          & Road       & 0.277  & \textbf{-0.070} & 0.274  & \textbf{-0.074} \\
    &          & Grass      & 0.109  & \textbf{-0.238} & 0.104  & \textbf{-0.244} \\
    & \CarRed  & Benign     & 0.436  &                & 0.439  &                \\
    &          & Road       & 0.228  & \textbf{-0.208} & 0.226  & \textbf{-0.213} \\
    &          & Grass      & 0.198  & \textbf{-0.238} & 0.195  & \textbf{-0.244} \\
    & \CarGray & Benign     & 0.347  &                & 0.341  &                \\
    &          & Road       & 0.238  & \textbf{-0.109} & 0.232  & \textbf{-0.109} \\
    &          & Grass      & 0.178  & \textbf{-0.169} & 0.177  & \textbf{-0.164} \\
    \bottomrule
  \end{tabular}
  \caption{AP@0.5 / AR@0.5 for benign vs.\ road- and grass-camouflage appearance across detectors for the overhead-based vehicle scenario.}
  \label{tab:car-ap-ar}
\end{table}
\clearpage
\endgroup

\begin{table}[H]
  \centering
  \tiny
  \caption{AP@0.5 / AR@0.5 for benign vs. adversarial road-camouflage on a single object (Stop Sign) across detectors}
  \label{tab:stop-sign-ap-ar}
  \begin{tabular}{lclcccr}
    \toprule
    \textbf{Model} & \textbf{Object} & \textbf{Appearance} & \textbf{AP@0.5} & $\boldsymbol{\Delta}$\textbf{AP} & \textbf{AR@0.5} & $\boldsymbol{\Delta}$\textbf{AR} \\
    \midrule
    \textbf{YOLOv3}
    & \TikZStop  & Benign    & 0.851   &            & 0.854   &            \\
    &            & Clock       & 0.446   & \textbf{-0.405}      & 0.444   & \textbf{-0.410}      \\
    &            & Soccer     & 0.683   & \textbf{-0.168} & 0.688   & \textbf{-0.166} \\
    \midrule
    \textbf{YOLOv5}
    & \TikZStop  & Benign    & 0.832   &            & 0.833   &            \\
    &            & Clock       & 0.436   & \textbf{-0.396}      & 0.431   & \textbf{-0.402}      \\
    &            & Soccer     & 0.505   & \textbf{-0.327} & 0.500   & \textbf{-0.333} \\
    \midrule
    \textbf{YOLOv8}
    & \TikZStop  & Benign    & 0.851   &            & 0.854   &            \\
    &            & Clock       & 0.356   & \textbf{-0.495}      & 0.358   & \textbf{-0.496}      \\
    &            & Soccer     & 0.436   & \textbf{-0.415} & 0.436   & \textbf{-0.418} \\
    \midrule
    \textbf{YOLOv11}
    & \TikZStop  & Benign    & 0.822   &            & 0.820   &            \\
    &            & Clock       & 0.307   & \textbf{-0.515}      & 0.302   & \textbf{-0.518}      \\
    &            & Soccer     & 0.396   & \textbf{-0.426} & 0.396   & \textbf{-0.424} \\
    \midrule
    \textbf{FRCNN}
    & \TikZStop  & Benign    & 0.723   &            & 0.729   &            \\
    &            & Clock       & 0.119   & \textbf{-0.604}      & 0.118   & \textbf{-0.611}      \\
    &            & Soccer     & 0.386   & \textbf{-0.337} & 0.389   & \textbf{-0.340} \\
    \midrule
    \textbf{DETR}
    & \TikZStop  & Benign    & 0.881   &            & 0.889   &            \\
    &            & Clock       & 0.416   & \textbf{-0.465}      & 0.417   & \textbf{-0.472}      \\
    &            & Soccer     & 0.545   & \textbf{-0.336} & 0.549   & \textbf{-0.340} \\
    \bottomrule
  \end{tabular}
\end{table}

\begin{table}[H]
  \centering
  \tiny
  \caption{AP@0.5 / AR@0.5 for benign vs.\ adversarial on overhead blue-car scenario across detectors.}
  \label{tab:bluecar-ap-ar}
  \begin{tabular}{lllcccr}
    \toprule
    \textbf{Model} & \textbf{Object} & \textbf{Condition} & \textbf{AP@0.5} & $\boldsymbol{\Delta}$\textbf{AP} & \textbf{AR@0.5} & $\boldsymbol{\Delta}$\textbf{AR} \\
    \midrule
    \textbf{YOLOv3}
    & \CarBlue   & Benign    & 0.822   &            & 0.822   &             \\
    &            & Adv       & 0.703   & \textbf{-0.119} & 0.707   & \textbf{-0.115} \\
    \midrule
    \textbf{YOLOv5}
    & \CarBlue   & Benign    & 0.713   &            & 0.714   &             \\
    &            & Adv       & 0.535   & \textbf{-0.178} & 0.539   & \textbf{-0.175} \\
    \midrule
    \textbf{YOLOv8}
    & \CarBlue   & Benign    & 0.436   &            & 0.430   &             \\
    &            & Adv       & 0.287   & \textbf{-0.149} & 0.285   & \textbf{-0.145} \\
    \midrule
    \textbf{YOLOv11}
    & \CarBlue   & Benign    & 0.634   &            & 0.640   &             \\
    &            & Adv       & 0.436   & \textbf{-0.198} & 0.438   & \textbf{-0.202} \\
    \midrule
    \textbf{FRCNN}
    & \CarBlue   & Benign    & 0.406   &            & 0.409   &             \\
    &            & Adv       & 0.010   & \textbf{-0.396} & 0.006   & \textbf{-0.403} \\
    \midrule
    \textbf{DETR}
    & \CarBlue   & Benign    & 0.297   &            & 0.293   &             \\
    &            & Adv       & 0.139   & \textbf{-0.158} & 0.134   & \textbf{-0.159} \\
    \bottomrule
  \end{tabular}
\end{table}

\clearpage
\begingroup
\raggedbottom
\subsection{Defense recovery detail}
Table~\ref{tab:defensesplat_with_benign_full} expands the defense evasion summary of main-body Table~\ref{tab:defensesplat_with_benign} into separate DefenseSplat (DS), DefenseSplat-RELU (DS-ReLU), and DEGauss columns; see caption for how its absolute point recovery values relate to the gap normalized recovery reported in the body.

\begin{table}[H]
  \centering
  \scriptsize
  \begin{tabular}{llrrrl}
    \toprule
    & &
    \multicolumn{3}{c}{\textbf{Detection rates (\%)}} & \\
    \cmidrule(lr){3-5}
    \textbf{Scene} & \textbf{Attack target} &
    \textbf{Benign} $\uparrow$ &
    \textbf{Attacked} $\downarrow$ &
    \textbf{Defended;} $\boldsymbol{\Delta}$ $\uparrow$ &
    \textbf{Defense} \\
    \midrule
    Car & Grass & 52.8 & 24.8 & 13.3 \textcolor{red}{(-11.5)} & DS \\
    &       &      &      & 13.1 \textcolor{red}{(-11.7)} & DS-ReLU \\
    &       &      &      & 0.1 \textcolor{red}{(-24.7)} & DEGauss \\
    \midrule
    Car & Road & 52.8 & 28.7 & 31.8 \textcolor{green!50!black}{(+3.1)} & DS \\
    &      &      &      & 31.3 \textcolor{green!50!black}{(+2.6)} & DS-ReLU \\
    &       &      &      & 3.6 \textcolor{red}{(-25.1)} & DEGauss \\
    \midrule
    Stop sign & Clock & 82.5 & 26.3 & 32.6 \textcolor{green!50!black}{(+6.3)} & DS \\
    &       &      &      & 32.8 \textcolor{green!50!black}{(+6.5)} & DS-ReLU \\
    &       &      &      & 42.6 \textcolor{green!50!black}{(+16.3)} & DEGauss \\
    \midrule
    Stop sign & Sports ball & 82.5 & 28.8 & 47.3 \textcolor{green!50!black}{(+18.5)} & DS \\
    &             &      &      & 46.1 \textcolor{green!50!black}{(+17.3)} & DS-ReLU \\
    &             &      &      & 57.1 \textcolor{green!50!black}{(+28.3)} & DEGauss \\
    \midrule
    Chair & Flooring & 82.4 & 10.5 & 13.1 \textcolor{green!50!black}{(+2.6)} & DS \\
    &               &      &      & 12.5 \textcolor{green!50!black}{(+2.0)} & DS-ReLU \\
    &               &      &      & 13.6 \textcolor{green!50!black}{(+3.1)} & DEGauss \\
    \midrule
    Horse & Zebra & 64.4 & 34.6 & 33.2 \textcolor{red}{(-1.4)} & DS \\
    &       &      &      & 31.9 \textcolor{red}{(-2.7)} & DS-ReLU \\
    &       &      &      & 42.9 \textcolor{green!50!black}{(+8.3)} & DEGauss \\
    \midrule
    Overall & -- & 70.1 & 25.6 & \textbf{28.6 (+3.0)} & DS \\
    &    &      &      & \textbf{28.9 (+2.3)} & DS-ReLU \\
    &    &      &      & \textbf{26.7 (+1.0)} & DEGauss \\
    \bottomrule
  \end{tabular}
  \caption{
    Defense-Splat (DS) recovery of intended-object detections after attack, expanded into the DS, DS-ReLU, and DEGauss variants summarized in main-body Table~\ref{tab:defensesplat_with_benign}.
    Detection rates report the mean percentage of views in which the intended object is detected, averaged across YOLOv\{3,5,8,11\}, Faster R-CNN, and DETR.
    The single DefenseSplat column in Table~\ref{tab:defensesplat_with_benign} corresponds to the DS/DS-ReLU variants broken out here, and the defended detection rates are consistent between the two tables.
    The two tables differ only in the parenthetical recovery metric: the values in parentheses here are \emph{absolute} changes in detection rate (defended $-$ attacked, in percentage points), whereas Table~\ref{tab:defensesplat_with_benign} reports recovery \emph{normalized} by the attack-induced gap, $(\text{defended}-\text{attacked})/(\text{benign}-\text{attacked})$.
    For example, Car$\to$Grass under DS appears as $-11.5$ points here and as $-41.4\%$ of the gap in Table~\ref{tab:defensesplat_with_benign}.
  }
  \label{tab:defensesplat_with_benign_full}
\end{table}

\clearpage
\subsection{Defense reproduction fidelity}
Tables~\ref{tab:defensesplat-reproduction-fidelity}--\ref{tab:degauss_reproduction_fidelity_three_scenes} validate our reimplementations of each defense against the metrics reported in their original papers to support validation claims in the Defense Evasion analysis (Table~\ref{tab:defensesplat_with_benign}) that evasion is not an artifact of a weakened defense.

\begin{table}[H]
  \centering
  \small
  \begin{tabular}{llrrrrr}
    \toprule
    \textbf{Source} & \textbf{Method} & $\boldsymbol{n}$ & $\boldsymbol{\Delta}$\textbf{\#G} $\downarrow$ & $\boldsymbol{\Delta}$\textbf{PSNR} $\uparrow$ & $\boldsymbol{\Delta}$\textbf{SSIM} $\uparrow$ & $\boldsymbol{\Delta}$\textbf{LPIPS} $\downarrow$ \\
    \midrule
    Paper Tanks\&Temples    & Ours      & paper & 54.4\% & +0.86 & +0.153 & -0.081 \\
    Paper Tanks\&Temples    & Ours+ReLU & paper & 55.8\% & +1.00 & +0.162 & -0.084 \\
    Ours T\&T (Truck/Train) & Ours      & 1/2   & 54.4\% & +1.08 & +0.220 & -0.127 \\
    Ours T\&T (Truck/Train) & Ours+ReLU & 1/2   & 55.2\% & +1.11 & +0.223 & -0.126 \\
    Paper Mip-NeRF 360      & Ours      & paper & 62.1\% & +2.14 & +0.161 & -0.101 \\
    Paper Mip-NeRF 360      & Ours+ReLU & paper & 63.5\% & +2.29 & +0.167 & -0.106 \\
    Ours Mip-NeRF (Bonsai)  & Ours      & 1/1   & 83.3\% & +3.33 & +0.250 & -0.199 \\
    Ours Mip-NeRF (Bonsai)  & Ours+ReLU & 1/1   & 83.7\% & +3.32 & +0.252 & -0.203 \\
    \bottomrule
  \end{tabular}
  \caption{DefenseSplat reproduction fidelity against Poison-Splat ($\epsilon=16/255$). Values are changes relative to vanilla 3DGS trained on poisoned images; $\Delta$\#G is the percentage reduction in final Gaussian count. In the $n$ column, ``paper'' denotes values quoted from the original DefenseSplat paper, and fractions (e.g.\ 1/2) denote scenes with complete metrics among those we trained locally. Our reproduction recovers the reported filtering effect, and our local runs show comparable or stronger reduction, consistent with the defense operating as intended.}
  \label{tab:defensesplat-reproduction-fidelity}
\end{table}

\begin{table}[H]
  \centering
  \small
  \begin{tabular}{llrrr}
    \toprule
    \textbf{Scene} & \textbf{Metric} & \textbf{Paper} & \textbf{Ours} & $\boldsymbol{\Delta}$ \\
    \midrule
    Bicycle & PSNR $\uparrow$   & 32.84  & 32.83  & -0.01 \\
    Bicycle & CLIP $\downarrow$ & 0.9358 & 0.9503 & +0.0145 \\
    Bicycle & CLIP-T $\downarrow$ & 0.2336 & 0.1963 & -0.0373 \\
    Bicycle & CLIP-D $\downarrow$ & 0.0535 & 0.1166 & +0.0631 \\
    \midrule
    Bear & PSNR $\uparrow$     & 32.20  & 32.13  & -0.07 \\
    Bear & CLIP $\downarrow$   & 0.8949 & 0.9370 & +0.0421 \\
    Bear & CLIP-T $\downarrow$ & 0.2940 & 0.2247 & -0.0693 \\
    Bear & CLIP-D $\downarrow$ & 0.0256 & 0.0101 & -0.0155 \\
    \midrule
    Face & PSNR $\uparrow$     & 33.92  & 33.91  & -0.01 \\
    Face & CLIP $\downarrow$   & 0.8860 & 0.9696 & +0.0836 \\
    Face & CLIP-T $\downarrow$ & 0.2193 & 0.2356 & +0.0163 \\
    Face & CLIP-D $\downarrow$ & 0.0325 & 0.1244 & +0.0919 \\
    \bottomrule
  \end{tabular}
  \caption{DEGauss reproduction fidelity on paper scenes using the paper's GaussianEditor edit settings, comparing reported paper metrics against our independent reimplementation. PSNR is a content-preservation metric (higher is better); CLIP, CLIP-T, and CLIP-D measure editing disruption (lower is better). Our reimplementation reproduces the reported preservation closely (PSNR within $0.07$ across scenes) and disrupts editing in the same regime, while being no stronger than the original; ComplicitSplat's evasion is therefore not an artifact of a weakened defense.
  The Face scene shows the largest CLIP/CLIP-D divergence; as the hardest preservation case, our reimplementation is toward content preservation over edit disruption.}
  \label{tab:degauss_reproduction_fidelity_three_scenes}
\end{table}
\clearpage
\endgroup
\end{document}